\documentclass[journal,twoside,web]{ieeecolor}
\usepackage{jsen}
\usepackage{cite}
\usepackage{amsmath,amssymb,amsfonts}
\usepackage{algorithmic}
\usepackage{graphicx}
\usepackage{textcomp}

\usepackage{booktabs}  \usepackage{algorithm}
\usepackage{cite} 

\usepackage{hyperref}
\hypersetup{
	hypertex=true, 
	colorlinks=true, 
	linkcolor=blue, 
	anchorcolor=blue, 
	citecolor=blue
}

\usepackage{amsmath,amsfonts,bm}

\def\eqref#1{equation~\ref{#1}}

\def\1{\bm{1}}

\DeclareMathAlphabet{\mathsfit}{\encodingdefault}{\sfdefault}{m}{sl}
\SetMathAlphabet{\mathsfit}{bold}{\encodingdefault}{\sfdefault}{bx}{n}

\usepackage{graphics}
\usepackage{marvosym}
\usepackage{bm}
\usepackage{ulem}
\usepackage[switch]{lineno}

\usepackage{url}

\usepackage{subcaption}
\usepackage{multirow}
\usepackage{color,xcolor}
\usepackage{colortbl}

\usepackage{adjustbox}

\usepackage {diagbox}
\usepackage{wasysym}
\usepackage{makecell}
\usepackage{caption}

\usepackage{mathrsfs}

\usepackage{tikz}
\usepackage{graphicx}
\usetikzlibrary{shapes.geometric, arrows}

\def\BibTeX{{\rm B\kern-.05em{\sc i\kern-.025em b}\kern-.08em
		T\kern-.1667em\lower.7ex\hbox{E}\kern-.125emX}}
\definecolor{abstractbg}{rgb}{0.89804,0.94510,0.83137}
\begin{document}
\title{VitalBench: A Rigorous Multi-Center Benchmark for Long-Term Vital Sign Prediction in Intraoperative Care}

\author{Xiuding Cai, Xueyao Wang, Sen Wang, Yaoyao Zhu, Jiao Chen and Yu Yao
	\thanks{This work was supported in part by the Sichuan Provincial Science and Technology Department under Grant 2022YFS0384 and 2022YFQ0108, and in part by the Chengdu Science and Technology Program under Grant 2022-YF04-00078-JH. (Corresponding authors: Jiao Chen and Yu Yao.)}
	\thanks{Xiuding Cai, Xueyao Wang, Sen Wang, Yaoyao Zhu and Yu Yao are with Chengdu Institute of Computer Application, Chinese Academy of Sciences, Chengdu, China, and with the School of Computer Science and Technology, University of Chinese Academy of Sciences, Beijing, China (e-mail: \{caixiuding20, wangxueyao221\}@mails.ucas.ac.cn, casitmed2022@163.com). }
	\thanks{Sen Wang was with Digital Intelligent Technology Branch, Southwest Oil \& Gas Field, PetroChina (e-mail: wangsen05@petrochina.com.cn).}
	\thanks{Yaoyao Zhu was with China Zhenhua Research Institute Co., Ltd., Guiyang, China (e-mail: zhuyaoyao19@mails.ucas.ac.cn).}
	\thanks{Jiao Chen was with Department of Anesthesiology, West China Hospital, Sichuan University \& The Research Units of West China (2018RU012) (e-mail: chenjiao@wchscu.cn).}}

\maketitle

\begin{abstract}
Intraoperative monitoring and prediction of vital signs are critical for ensuring patient safety and improving surgical outcomes. Despite recent advances in deep learning models for medical time-series forecasting, several challenges persist, including the lack of standardized benchmarks, incomplete data, and limited cross-center validation. To address these challenges, we introduce VitalBench, a novel benchmark specifically designed for intraoperative vital sign prediction. VitalBench includes data from over 4,000 surgeries across two independent medical centers, offering three evaluation tracks: complete data, incomplete data, and cross-center generalization. This framework reflects the real-world complexities of clinical practice, minimizing reliance on extensive preprocessing and incorporating masked loss techniques for robust and unbiased model evaluation. By providing a standardized and unified platform for model development and comparison, VitalBench enables researchers to focus on architectural innovation while ensuring consistency in data handling. This work lays the foundation for advancing predictive models for intraoperative vital sign forecasting, ensuring that these models are not only accurate but also robust and adaptable across diverse clinical environments. Our code and data are available at \url{https://github.com/XiudingCai/VitalBench}.
\end{abstract}


\section{Introduction}

\IEEEPARstart{I}{ntraoperative} monitoring and prediction of vital signs, such as heart rate, blood pressure, and oxygen saturation, are critical for ensuring patient safety and optimizing surgical outcomes~\cite{kumar2020continuous, da2018internet, kaieski2020application, ahmed2023machine}. These physiological indicators provide real-time insight into a patient's physiological state and enable early detection of potential complications. Accurate forecasting of vital sign trend allows surgical teams to anticipate risks and implement timely interventions, thereby reducing adverse events and improving postoperative recovery~\cite{kyriacos2011monitoring, feng2021predicting, kristinsson2022prediction, dubatovka2024predicting}.

Recent advances in deep learning~\cite{sun2020review, sun2023few, jarrett2023clairvoyance, moor2023predicting, zhang2024imbalanced}, particularly in managing complex, multivariate time-series data, have shown considerable promise in medical data analysis, pattern recognition, and improving predictive accuracy. Models such as Recurrent Neural Networks (RNNs)~\cite{salinas2020deepar, NIPS2015_07563a3f, lai2018modeling}, Convolutional Neural Networks (CNNs)~\cite{bai2018empirical, wu2023timesnet, wang2023micn}, Multi-Layer Perceptrons (MLPs)~\cite{Zeng2023Dlinear, zhang2022less}, Graph Neural Networks (GNNs)~\cite{wu2020mtgnn, yi2024fouriergnn} and Transformers~\cite{zhou2021informer, zhou2022fedformer, 10906418, 11194085} have achieved notable success across a range of healthcare applications, providing novel methodologies for vital sign prediction. These approaches are capable of identifying latent patterns within large-scale medical datasets, thereby significantly enhancing the precision and reliability of clinical predictions.

Despite the promising results of deep learning in vital sign prediction, several challenges persist:

\textbf{Challenge 1: Lack of Standardized Benchmarks and Inconsistent Preprocessing.}
The field of vital sign prediction is characterized by a fragmented landscape where studies often rely on {proprietary datasets and diverse preprocessing pipelines}~\cite{lee2021deep, chen2022nonlinear}. This fragmentation not only obstructs direct model comparisons but also undermines reproducibility. Without a unified benchmark, the community faces difficulties in identifying fundamental improvements or addressing gaps across different approaches, stalling collective advancements.

\textbf{Challenge 2: Assumption of Complete Data Availability.}
Many existing studies implicitly assume complete vital sign data~\cite{bahador2021multimodal, he2023transformer}, where all monitoring variables are consistently available for each patient. In practice, however, sensor failures, detachment events, and individualized monitoring setups result in frequent and structured missingness. Models trained under this assumption often degrade in performance when facing incomplete inputs, emphasizing the need for methods that explicitly accommodate heterogeneous data availability.

\textbf{Challenge 3: Single-Center Validation.}
Predominantly, current studies rely on data from a single medical center~\cite{yeche2021hirid, rockenschaub2024impact}. However, variability in patient demographics, sensor devices, and clinical protocols across institutions creates a fundamental barrier to real-world applicability. Without cross-center validation, the robustness and transferability of these models remain untested, raising concerns about their reliability and generalizability in broader clinical deployment.

To tackle these challenges, we propose VitalBench, a comprehensive, clinically grounded benchmark for long-term intraoperative vital sign forecasting. VitalBench comprises data from 4,183 surgeries collected across two medical centers, with 962 from VitalDB~\cite{lee2022vitaldb} and 3,221 from MOVER-SIS~\cite{samad2023medical}. To promote progress in addressing real-world challenges, VitalBench introduces three targeted evaluation tracks:

\textbf{Track 1: Complete Variables Data.}
An idealized track where all vital sign variables are fully observed. To preserve clinical realism, raw data are used without imputation, as synthetic filling may distort physiological dynamics. Masked loss functions mitigate the impact of missing values, enabling fair and robust model comparison under consistent conditions.

\textbf{Track 2: Incomplete Variables Data.}
A clinically grounded setting where subsets of vital sign variables are missing due to sensor dropout or case-specific configurations. This track challenges models to handle variable missingness and promotes the development of architectures inherently robust to incomplete heterogeneous inputs.

\textbf{Track 3: Cross-Center Generalization.}
A transferability evaluation across institutions, where models trained on one medical center are tested on another. It examines robustness to distributional shifts caused by differences in patient populations, monitoring devices, and surgical protocols, encouraging generalizable and deployable forecasting solutions.

Using {VitalBench}, we conduct a systematic evaluation of a wide range of state-of-the-art time-series forecasting models, including RNNs, CNNs, MLPs, GNNs, and Transformers. The results yield several key insights into the core challenges and design principles of intraoperative vital sign prediction:
\begin{enumerate}\item \textbf{Modeling inter-variable dependencies is crucial for reliable forecasting.} Architectures that explicitly mix variables consistently outperform variable-independent ones, particularly under incomplete inputs or heterogeneous data. This reflects the physiological coupling of vital signs and underscores the need to capture cross-variable dynamics for robust and scalable modeling.
	
	\item \textbf{Modeling missingness directly outperforms imputation.} models trained with proposed masked loss outperform those using conventional imputation, highlighting the value of preserving the natural missingness patterns in clinical data rather than introducing potentially spurious or temporally inconsistent values that could mislead predictions and reduce clinical reliability.
	
	\item \textbf{Data scale enhances generalization, yet cross-center gaps persist.} 
	Training on larger datasets consistently enhances performance, and in some prediction horizons, models trained on large cohorts match results on smaller cohorts, indicating feasible cross-domain transfer. Yet, performance drops remain across centers in most conditions, highlighting the clinical generalization challenge.

\end{enumerate}

\textbf{Our main contributions are as follows:}

\begin{enumerate}
	\item We introduce {VitalBench}, the first multi-center benchmark for long-term intraoperative vital sign forecasting, with three clinically relevant tracks—complete data, incomplete data, and cross-center generalization—designed to evaluate model performance and generalization under real-world surgical conditions.
	
	\item We design an {end-to-end evaluation framework} that reflects real-world clinical challenges, minimizing reliance on heavy preprocessing and employing masked loss to handle missing data, thereby  ensuring consistent, reproducible, and fair model comparisons.
	
	\item Using VitalBench, we conduct a {comprehensive empirical study} uncovering key insights for vital sign forecasting, highlighting the need for models that capture inter-variable dependencies, handle variable channel dimensions, and generalize across clinical centers.
	
\end{enumerate}
 \section{Related Work}

\begin{table*}[htbp]
	\fontsize{10pt}{12pt}\selectfont
	\centering
	\renewcommand{\arraystretch}{1.}
	\caption{Summary of statistics for different time-series datasets. \#Instance represents the number of instances, and MC denotes whether the dataset is multi-center.}
	\label{table:datasets}
	\begin{adjustbox}{max width=0.8\linewidth}
		\begin{tabular}{lllllrcc}
			\hline 
			Dataset & Domain & Frequency & Lengths & Variables & Volume & \#Instance & MC \\
			\hline 
			Exchange~\cite{lai2018modeling} & Exchange & 1 day & 7,588 & 8 &623 {KB} & 1 &  \\
			Electricity~\cite{misc_electricityloaddiagrams20112014_321} & Electricity & 1 hour & 26,304 & 321 & 92 {MB}& 1 &  \\
			ETTm2~\cite{zhou2021informer} & Electricity & 15 mins & 57,600 & 7 &9.3 {MB} & 1 &  \\
			Traffic~\cite{wu2021autoformer} & Traffic & 1 hour & 17,544 & 862 & 131 {MB} & 1 &     \\

			Solar~\cite{lai2018modeling} & Weather & 10 mins & 52,560 & 137 & 8.3 {MB} & 1 &    \\ 
Weather~\cite{wu2021autoformer} & Weather & 10 mins & 52,696 & 21 &7.0 {MB} & 1 &    \\
			\midrule
ILI~\cite{wu2021autoformer} & Health & 1 day & 966 & 7 &67.6 {KB} & 1 &   \\
			Covid-19~\cite{panagopoulos2021transfer} & Health & 1 day & 1,392 & 948 &5.0 {MB} & 1 &   \\
\midrule
\rowcolor[RGB]{247,225,237} 
VitalBench (Ours) &  &  &  &  & 488.7 {MB} & 4,183 & \checkmark \\
			\rowcolor{pink!25} \quad --~VitalDB & Health & 5 secs & 2,697,323 & 17 &458.0 {MB}& 962 &  \\
			\rowcolor{blue!10} \quad --~MOVER & Health & 1 min & 817,161 & 33 & 130.7 {MB}& 3,221 &  \\
			\hline
		\end{tabular}
	\end{adjustbox}
\end{table*} 
\subsection{Time Series Forecasting}
Time series forecasting has progressed from classical statistical methods~\cite{box1970distribution} to deep learning techniques capable of capturing complex temporal dependencies~\cite{zhou2021informer,Zeng2023Dlinear,nie2023a,cai2024mambats,10908642,zhang2024learning}, driven by the growth of large-scale datasets and the demand for models handling intricate patterns.

RNNs~\cite{salinas2020deepar,NIPS2015_07563a3f,lai2018modeling} pioneered long-term sequence forecasting (LTSF) by modeling sequential dependencies. Temporal Convolutional Networks (TCNs)~\cite{bai2018empirical, wu2023timesnet} emerged as an alternative, effectively capturing local temporal patterns with their localized receptive fields, whileCollectively, these benchmarks enhance reproducibility in ICU research yet fall short of capturing the real-time, high-frequency, and multi-center nature of intraoperative vital-sign prediction. Multi-Scale Isometric Convolution Network (MICN) extended temporal modeling by integrating local and global dependencies across scales, enhancing performance in diverse tasks~\cite{wang2023micn}.

Transformers~\cite{vaswani2017attention}, initially designed for natural language processing, have gained prominence in time series forecasting due to their self-attention mechanism, which excels at capturing long-range dependencies~\cite{zhou2021informer, zhou2022fedformer, 10669781}. However, the quadratic complexity of self-attention remains a computational bottleneck. Recent innovations, such as patch-based approaches like PatchTST~\cite{nie2023a}, address this limitation, making Transformers more practical for large-scale applications~\cite{liu2024itransformer}.

MLPs, have also shown promise in time series tasks, offering simplicity and efficiency in specific scenarios~\cite{Zeng2023Dlinear, zhang2022less}. GNNs~\cite{10477486}, like MTGNN~\cite{wu2020mtgnn}, have been particularly effective in multivariate time series, combining graph convolutional and temporal layers to model spatial-temporal relationships. FourierGNN~\cite{yi2024fouriergnn} further explores this domain by treating the entire time series as a hypervariate graph and leveraging Fourier transforms for global dependency modeling.

Recently, state-space models~\cite{fu2022h3, gu2021efficiently, zhang2023spacetime, gu2023mamba} have been introduced to time series forecasting, offering a fresh perspective. For instance, MambaTS~\cite{cai2024mambats} achieves linear complexity while effectively modeling global dependencies, demonstrating superior performance compared to Transformer-based methods.

\subsection{Perioperative Time Series Forecasting}

Recent years have seen notable progress in applying deep learning to intraoperative vital sign forecasting. Lee et al.~\cite{lee2021deep} developed a long short-term memory (LSTM) model to predict the bispectral index during total intravenous anesthesia using drug infusion histories and demographic features. He et al.~\cite{he2023transformer} further enhanced this framework by combining an LSTM with a gated residual network (GRN) for improved feature extraction. Bahador et al.~\cite{bahador2021multimodal} applied deep learning to EEG-based depth-of-anesthesia estimation, demonstrating the capacity of neural models to capture complex physiological dynamics.

Despite recent progress, existing research remains fragmented and limited. Many studies rely on proprietary datasets or closed-source implementations, impeding reproducibility and external validation. Even public datasets suffer from heterogeneous preprocessing and task definitions, hindering consistent evaluation. Most efforts focus on short-term or single-variable predictions within single-center cohorts, overlooking long-term dependencies, inter-variable correlations, and cross-hospital generalization. As a result, the field lacks a standardized, openly accessible benchmark for fair, reproducible, and clinically relevant assessment of forecasting models.

\subsection{Benchmarks}

\begin{figure*}[t]
	\centering
	\includegraphics[width=1\linewidth]{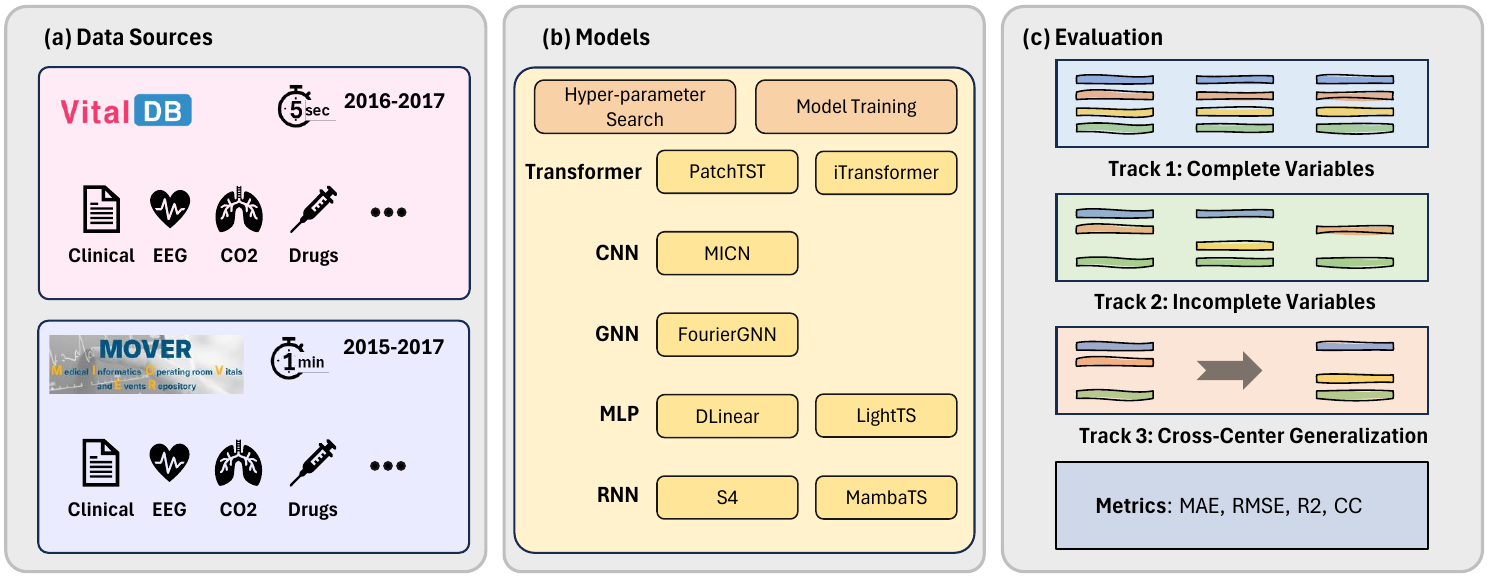}
	\caption{Overview of the VitalBench Workflow.}
	\label{fig:pipeline}
\end{figure*}

Benchmarks drive scientific progress by offering standardized datasets, metrics, and protocols for fair and reproducible model comparison~\cite{saporta2022benchmarking, ektefaie2024evaluating}. LTSF benchmarks such as Libra~\cite{bauer2021libra}, TSLib~\cite{wu2023timesnet}, and TFB~\cite{qiu2024tfb} facilitate cross-domain evaluation and promote methodological advances through unified training and validation protocols.

Medical forecasting, however, faces additional complexities beyond temporal generalization. Clinical data are heterogeneous, sporadic, noisy, and patient-dependent, collected from multiple devices under dynamic physiological and treatment conditions. Consequently, medical benchmarks must address missingness, label imbalance, and inter-patient variability, which are rarely emphasized in standard LTSF frameworks.

Several works have standardized clinical time-series evaluation. Harutyunyan et al.~\cite{harutyunyan2019multitask} proposed four MIMIC-III~\cite{johnson2016mimic} benchmarks for mortality, length-of-stay, physiologic decline, and phenotype classification, exploring multitask learning and deep supervision. HiRID~\cite{yeche2021hirid} provides high-resolution ICU data with structured tasks but sparse interventions and gradually evolving trajectories. Van De Water et al.~\cite{van2023yet} introduced YAIB, supporting multiple ICU datasets and predefined tasks including mortality, sepsis, and kidney function, yet still focusing on long-term or static outcomes. Clairvoyance~\cite{jarrett2023clairvoyance} provides a unified pipeline for clinical time-series modeling and evaluation, not designed for vital-sign forecasting. Collectively, these benchmarks enhance reproducibility in ICU research yet fall short of capturing the real-time, high-frequency, and multi-center nature of intraoperative vital-sign prediction.

Intraoperative vital-sign forecasting presents distinct challenges. Signals are recorded at sub-minute resolution under rapidly changing physiological states and frequent interventions, leading to artifacts, missingness, and strong nonstationarity across patients and procedures. Moreover, multi-center validation remains limited~\cite{yeche2021hirid, rockenschaub2024impact}, despite its importance for generalization across populations and sites. Patient-level data splits are also crucial to prevent leakage and capture inter-patient variability. These gaps highlight the need for a dedicated benchmark for real-time intraoperative forecasting.

Table~\ref{table:datasets} summarizes existing time series datasets and introduces our proposed benchmark, {VitalBench}, specifically designed to address these limitations, providing a foundation for robust and generalizable perioperative vital-sign prediction.
 
\section{VitalBench}

\subsection{Task Description}

We consider the problem of multivariate time series forecasting in the operating room. Given a retrospective window $\mathbf{x} = {\mathbf{x}_1, \mathbf{x}_2, \cdots, \mathbf{x}_L}$ of length $L$ with $K$ variables, where $\mathbf{x}_t \in \mathbb{R}^K$ represents the values of the $K$ variables at time step $t$, the goal is to predict the values for the next $T$ time steps, denoted as $\mathbf{y} = {\mathbf{x}_{L+1}, \mathbf{x}_{L+2}, \cdots, \mathbf{x}_{L+T}}$. 

In this formulation, the \(K\) variables include both static covariates, such as patient demographics and procedure type, and dynamic variables, such as heart rate and blood pressure, which evolve over time. The forecasting task focuses exclusively on predicting the future values of the dynamic variables, while static covariates serve as conditioning context throughout the prediction horizon.

\subsection{Data Collection}

\textbf{VitalDB Dataset.}  
VitalDB~\cite{lee2022vitaldb} is a publicly available, high-resolution perioperative dataset from Seoul National University Hospital, comprising 6,388 non-cardiac surgeries performed between August 2016 and June 2017. Each case includes physiological waveforms and structured clinical data sampled at 5-second intervals. We accessed the data via the official API and followed the format of Lee et al.~\cite{lee2021deep} to extract key variables, including demographics, anesthetic parameters, and vital signs, resulting in one wide-format CSV file per surgery, with rows as time steps and columns as variables. After filtering for data quality and completeness while de-identifying patient information, we retained 962 surgeries. See Appendix Table~\ref{tab:variables_vitaldb} for variable details.

\textbf{MOVER-SIS Dataset.}  
The MOVER dataset, currently the largest open perioperative dataset, includes approximately 59,000 patients' surgical data from around 83,000 surgeries conducted between 2015 and 2022 at the University of California, Irvine Medical Center. The dataset is split into two parts: the SIS (Surgical Information Systems, USA) dataset and the EPIC (Epic Systems, USA) dataset. Given the absence of ventilator parameters in the EPIC dataset, we focused solely on the SIS dataset to align with the variables selected from VitalDB. We extracted a subset from nine long-format tables, reformatted them into wide-format CSV files (one per surgery), and harmonized them with the VitalDB structure. Selected variables include demographics, anesthetic agents, ventilator parameters, and vital signs. After preprocessing and de-identification, 3,221 surgeries were included for analysis. See Appendix Table~\ref{tab:variables_mover} for full variable list.

\subsection{Data Preprocessing}

\begin{figure*}[!t]
	\centering
	\includegraphics[width=0.86\linewidth]{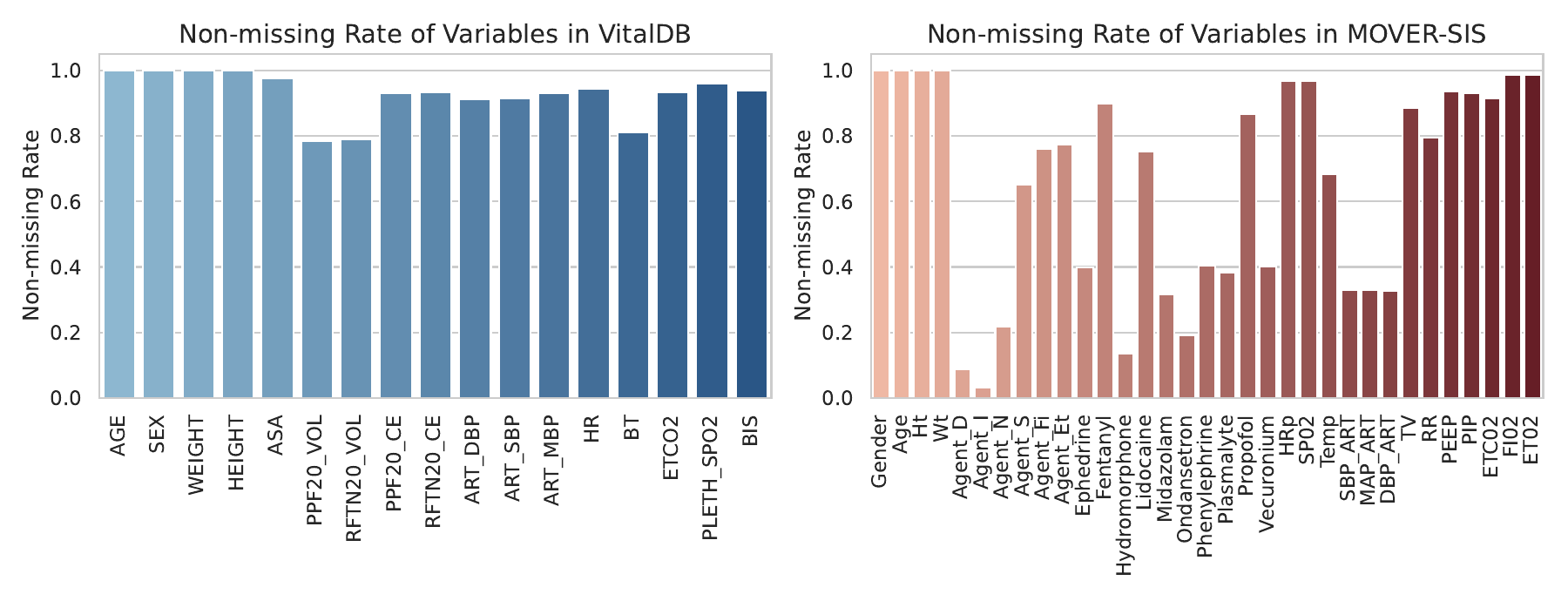}
	\caption{Non-missing rate of variables across different datasets: VitalDB (left) and MOVER-SIS (right).}
	\label{fig:missing_rate}
\end{figure*}

For each surgical case, we generated input–target pairs by sliding a temporal window across the intraoperative record. Each pair consists of an input sequence $\mathbf{x} \in \mathbb{R}^{L \times K}$ and a target sequence $\mathbf{y} \in \mathbb{R}^{T \times K_d}$, where $L$ denotes the retrospective window length, $T$ the prediction horizon, $K$ the total number of variables, and $K_d \leq K$ the subset of dynamic variables to be forecasted. To handle cold-start conditions, zero-padding was applied to initial steps, but no further preprocessing (e.g., imputation) was used. As shown in Figure~\ref{fig:missing_rate}, VitalDB and MOVER-SIS exhibit substantial missingness (20-80\%) that reflects real clinical dynamics. Imputation could distort these properties; thus, we retained missing values to preserve data fidelity and encourage learning under realistic conditions. The effectiveness of this design is validated in Section~\ref{sec:ablation}.

\subsection{Masked Loss Function}

Since no imputation was performed, we introduced a masked loss function to prevent zero-filled entries from biasing model optimization. Specifically, we applied Mean Squared Error (MSE) as the base objective and applied a binary mask to exclude missing labels from gradient computation:
\begin{equation}
	\mathcal{L} = \mathbb{E}_{(\mathbf{x}, \mathbf{y})} \left[ \frac{1}{T \cdot K_d} \sum_{t=1}^T \sum_{k=1}^{K_d} \mathbf{m}_{t,k} \left(\hat{y}_{t,k} - y_{t,k}\right)^2 \right],
\end{equation}
where $\mathbf{m} \in {0,1}^{T \times K_d}$ is a binary mask indicating whether each target value is observed ($\mathbf{m}_{t,k}=1$) or missing ($\mathbf{m}_{t,k}=0$).

This masked formulation ensures that only valid supervision signals contribute to the loss, thereby avoiding distortions introduced by missing entries and maintaining data fidelity. 

\begin{table*}[!t]
	\fontsize{10pt}{12pt}\selectfont
	\centering
	\renewcommand{\arraystretch}{1.1}
	\caption{Performance comparison of different models on the complete variables track. We report the average performance across different prediction horizons. For the full results, refer to Table~\ref{tab:main_track_full} in the Appendix.
	}
	\label{tab:main_track}
	\begin{adjustbox}{max width=\linewidth}
		\begin{tabular}{@{}cccccccccc@{}}
			\toprule
\multicolumn{2}{c}{{\diagbox[width=8em, height=3.9em]{Metrics}{Models}}}                                                   & {\begin{tabular}[c]{@{}c@{}}MambaTS\\ (2024)\end{tabular}} & {\begin{tabular}[c]{@{}c@{}}LightTS\\ (2023)\end{tabular}} & {\begin{tabular}[c]{@{}c@{}}iTransformer\\ (2024)\end{tabular}} & {\begin{tabular}[c]{@{}c@{}}FourierGNN\\ (2024)\end{tabular}} & {\begin{tabular}[c]{@{}c@{}}PatchTST\\ (2023)\end{tabular}} & {\begin{tabular}[c]{@{}c@{}}MICN\\ (2024)\end{tabular}} & {\begin{tabular}[c]{@{}c@{}}DLinear\\ (2023)\end{tabular}} & {\begin{tabular}[c]{@{}c@{}}S4\\ (2022)\end{tabular}} \\ 
			\midrule
\midrule

			\multicolumn{1}{c|}{\multirow{4}{*}{\rotatebox{90}{VitalDB}}} & \multicolumn{1}{c|}{MAE} & \cellcolor[RGB]{255,235,245} \textbf{0.034$\pm$0.011} &                                        0.038$\pm$0.011 &                     \cellcolor{blue!8} 0.037$\pm$0.010 &                                        0.040$\pm$0.013 &  0.059$\pm$0.064 &   0.045$\pm$0.014 &                     0.055$\pm$0.012 &               0.050$\pm$0.013 \\ 
			\multicolumn{1}{c|}{} & \multicolumn{1}{c|}{RMSE} & \cellcolor[RGB]{255,235,245} \textbf{0.057$\pm$0.015} &                                        0.061$\pm$0.015 &                     \cellcolor{blue!8} 0.059$\pm$0.015 &                                        0.064$\pm$0.019 &  0.128$\pm$0.067 &   0.087$\pm$0.025 &                     0.093$\pm$0.018 &               0.089$\pm$0.022 \\ 
			\multicolumn{1}{c|}{} & \multicolumn{1}{c|}{R2} & \cellcolor[RGB]{255,235,245} \textbf{0.950$\pm$0.032} &                                        0.942$\pm$0.034 &                     \cellcolor{blue!8} 0.947$\pm$0.034 &                                        0.937$\pm$0.045 &  0.594$\pm$2.777 &   0.863$\pm$0.077 &                     0.858$\pm$0.055 &               0.862$\pm$0.065 \\ 
			\multicolumn{1}{c|}{} & \multicolumn{1}{c|}{CC} & \cellcolor[RGB]{255,235,245} \textbf{0.862$\pm$0.167} &                                        0.850$\pm$0.168 &                     \cellcolor{blue!8} 0.862$\pm$0.167 &                                        0.836$\pm$0.190 &  0.787$\pm$0.232 &   0.831$\pm$0.166 &                     0.827$\pm$0.164 &  0.821$\pm$0.176 \\ \cline{1-10}

\midrule
			
\multicolumn{1}{c|}{\multirow{4}{*}{\rotatebox{90}{MOVER-SIS}}} & \multicolumn{1}{c|}{MAE} & \cellcolor[RGB]{255,235,245} \textbf{0.040$\pm$0.022} &                                        0.044$\pm$0.023 &                                        0.046$\pm$0.034 &                     \cellcolor{blue!8} 0.042$\pm$0.021 &  0.058$\pm$0.035 &   0.059$\pm$0.070 &                     0.074$\pm$0.065 &               0.067$\pm$0.067 \\ 
			\multicolumn{1}{c|}{} & \multicolumn{1}{c|}{RMSE} & \cellcolor{blue!8} 0.080$\pm$0.032 &                                        0.081$\pm$0.038 &                                        0.087$\pm$0.043 &  \cellcolor[RGB]{255,235,245} \textbf{0.079$\pm$0.034} &  0.127$\pm$0.045 &   0.137$\pm$0.074 &                     0.130$\pm$0.066 &               0.133$\pm$0.068 \\ 
			\multicolumn{1}{c|}{} & \multicolumn{1}{c|}{R2} & 0.805$\pm$1.155 &  \cellcolor[RGB]{255,235,245} \textbf{0.859$\pm$0.613} &                                        0.779$\pm$1.349 &                     \cellcolor{blue!8} 0.807$\pm$1.218 &  0.770$\pm$1.024 &   0.488$\pm$3.798 &                     0.620$\pm$2.353 &               0.599$\pm$2.634 \\ 
			\multicolumn{1}{c|}{} & \multicolumn{1}{c|}{CC} & \cellcolor[RGB]{255,235,245} \textbf{0.843$\pm$0.173} &                                        0.802$\pm$0.203 &                                        0.806$\pm$0.222 &                     \cellcolor{blue!8} 0.823$\pm$0.191 &  0.765$\pm$0.248 &   0.804$\pm$0.175 &                     0.781$\pm$0.187 &  0.770$\pm$0.240 \\ \cline{1-10}  
			
			\bottomrule
			
\end{tabular}
	\end{adjustbox}
\end{table*}
 
\subsection{Track Design}

To foster clinically reliable forecasting models, VitalBench introduces three progressively challenging tracks that reflect increasing realism in intraoperative data.

\noindent \textbf{Track 1: Complete Variables Data.} An idealized track where all physiological variables are fully observed, providing a benchmark for model performance under complete and uninterrupted intraoperative monitoring.

\noindent \textbf{Track 2: Incomplete Variables Data.} A realistic clinical setting where some variables are missing due to sensor dropout or incomplete documentation, requiring models to handle sparsity without imputation.

\noindent \textbf{Track 3: Cross-Center Generalization.} A transferability test in which models trained at one medical center are evaluated on another, assessing generalization across patient populations, surgical workflows, and monitoring systems.
 
\section{Experiments}

\subsection{Experimental Setup}

\noindent \textbf{Dataset Splitting.} The dataset was split based on the chronological order of surgeries, with a 7:1:2 ratio for training, validation, and test sets, ensuring temporal validity.

\noindent \textbf{Baselines and Metrics.} We selected a set of state-of-the-art (SOTA) time series forecasting models, representing both variable-independent approaches (e.g., iTransformer~\cite{liu2024itransformer}, LightTS~\cite{zhang2022less}, MICN~\cite{wang2023micn}, FourierGNN~\cite{yi2024fouriergnn}, MambaTS~\cite{cai2024mambats}) and variable-dependent models (e.g., DLinear~\cite{Zeng2023Dlinear}, PatchTST~\cite{nie2023a}, S4~\cite{gu2021efficiently}). Since Tracks 2 and 3 require handling varying input dimensions, models restricted to fixed feature sizes (e.g., LightTS, MICN) were excluded. For all experiments, we used standard evaluation metrics, including Mean Absolute Error (MAE), Root Mean Squared Error (RMSE), R-squared (R$^2$), and Correlation Coefficient (CC), to ensure a comprehensive and consistent assessment of model performance. Lower values for MAE and RMSE indicate better accuracy, while higher values for R$^2$ and CC suggest improved model fit and correlation.

\noindent \textbf{Track Design.} The VitalDB and MOVER-SIS datasets differ in sampling frequency (12 Hz vs. 1 Hz). Guided by clinical experts, we chose prediction horizons of 1, 3, 10, and 30 minutes, reflecting common clinical needs for intraoperative forecasting. Following Lee et al.~\cite{lee2021deep}, a 30-minute retrospective window was used for both datasets. For VitalDB, this corresponds to 360 input steps and 12, 36, 120, and 360 prediction steps, while for MOVER-SIS, 30 input steps and 1, 3, 10, and 30 prediction steps were used. These configurations were consistently applied across all three tracks, with VitalDB downsampled to 1 Hz in Track 3 to match MOVER-SIS. 

In Track 2 and 3, missing data were simulated by randomly dropping variable channels, including static covariates and dynamic physiological variables, while retaining at least one dynamic variable per patient. The same missingness pattern was applied to training, validation, and test to ensure consistent and fair comparisons across baselines.

\begin{table*}[!ht]
	\fontsize{10pt}{12pt}\selectfont
	\centering
	\renewcommand{\arraystretch}{1.1}
	\caption{Performance comparison of different models on the incomplete variables track. We report the average performance across different prediction horizons. For the full results, refer to Table~\ref{tab:main_masked_track_full} in the Appendix.
	}
	\label{tab:main_masked_track}
	\begin{adjustbox}{max width=0.78\linewidth}
		\begin{tabular}{@{}cccccccc@{}}
\cline{1-8}
			\multicolumn{2}{c}{{\diagbox[width=8em, height=3.9em]{Metrics}{Models}}}                                                   & {\begin{tabular}[c]{@{}c@{}}MambaTS\\ (2024)\end{tabular}} & {\begin{tabular}[c]{@{}c@{}}iTransformer\\ (2024)\end{tabular}} & {\begin{tabular}[c]{@{}c@{}}FourierGNN\\ (2024)\end{tabular}} & {\begin{tabular}[c]{@{}c@{}}PatchTST\\ (2023)\end{tabular}} &  {\begin{tabular}[c]{@{}c@{}}Dlinear\\ (2023)\end{tabular}} & {\begin{tabular}[c]{@{}c@{}}S4\\ (2022)\end{tabular}} \\ 
			\midrule
\midrule
\multicolumn{1}{c|}{\multirow{4}{*}{\rotatebox{90}{VitalDB}}} & \multicolumn{1}{c|}{MAE} & \cellcolor[RGB]{255,235,245} \textbf{0.034$\pm$0.004} &                     \cellcolor{blue!8} 0.038$\pm$0.003 &                                        0.039$\pm$0.003 &                     0.043$\pm$0.005 &                     0.053$\pm$0.004 &              0.049$\pm$0.005 \\ 
			\multicolumn{1}{c|}{} & \multicolumn{1}{c|}{RMSE} & \cellcolor[RGB]{255,235,245} \textbf{0.057$\pm$0.003} &                     \cellcolor{blue!8} 0.060$\pm$0.004 &                                        0.063$\pm$0.002 &                     0.081$\pm$0.008 &                     0.092$\pm$0.007 &              0.088$\pm$0.007 \\ 
			\multicolumn{1}{c|}{} & \multicolumn{1}{c|}{R2} & \cellcolor[RGB]{255,235,245} \textbf{0.864$\pm$0.156} &                     \cellcolor{blue!8} 0.852$\pm$0.150 &                                        0.841$\pm$0.154 &                     0.762$\pm$0.177 &                     0.704$\pm$0.179 &              0.734$\pm$0.160 \\ 
			\multicolumn{1}{c|}{} & \multicolumn{1}{c|}{CC} & \cellcolor[RGB]{255,235,245} \textbf{0.839$\pm$0.056} &                     \cellcolor{blue!8} 0.829$\pm$0.067 &                                        0.806$\pm$0.055 &                     0.810$\pm$0.045 &                     0.793$\pm$0.054 &  0.794$\pm$0.043 \\ \cline{1-8} 
			
\midrule
			
\multicolumn{1}{c|}{\multirow{4}{*}{\rotatebox{90}{MOVER-SIS}}} & \multicolumn{1}{c|}{MAE} & \cellcolor[RGB]{255,235,245} \textbf{0.048$\pm$0.005} &                     \cellcolor{blue!8} 0.052$\pm$0.002 &                                        0.057$\pm$0.006 &                     0.060$\pm$0.007 &                     0.073$\pm$0.003 &              0.065$\pm$0.006 \\ 
			\multicolumn{1}{c|}{} & \multicolumn{1}{c|}{RMSE} & \cellcolor[RGB]{255,235,245} \textbf{0.097$\pm$0.005} &                     \cellcolor{blue!8} 0.102$\pm$0.003 &                                        0.106$\pm$0.006 &                     0.126$\pm$0.005 &                     0.127$\pm$0.003 &              0.130$\pm$0.005 \\ 
			\multicolumn{1}{c|}{} & \multicolumn{1}{c|}{R2} & \cellcolor[RGB]{255,235,245} \textbf{0.835$\pm$0.254} &                                        0.820$\pm$0.355 &                     \cellcolor{blue!8} 0.829$\pm$0.154 &                     0.747$\pm$0.303 &                     0.750$\pm$0.275 &              0.730$\pm$0.344 \\ 
			\multicolumn{1}{c|}{} & \multicolumn{1}{c|}{CC} & \cellcolor[RGB]{255,235,245} \textbf{0.832$\pm$0.049} &                     \cellcolor{blue!8} 0.812$\pm$0.031 &                                        0.808$\pm$0.040 &                     0.787$\pm$0.034 &                     0.788$\pm$0.047 &  0.774$\pm$0.038 \\ \cline{1-8} 
			
\bottomrule
\end{tabular}
	\end{adjustbox}
\end{table*}

\begin{table*}[!ht]
	\fontsize{10pt}{12pt}\selectfont
	\centering
	\renewcommand{\arraystretch}{1.1}
	\caption{Performance comparison of different models on the cross-center generalization track. We report the average performance across different prediction horizons. For the full results, refer to Table~\ref{tab:main_dg_track_full} in the Appendix.
	}
	\label{tab:main_dg_track}
	\begin{adjustbox}{max width=0.9\linewidth}
		\begin{tabular}{@{}ccccccccc@{}}
\cline{1-9}
			\multicolumn{2}{c}{{\diagbox[width=8em, height=3.9em]{Metrics}{Models}}}                                                   & {\begin{tabular}[c]{@{}c@{}}\makecell[c]{Upper\\Bound} \\ \end{tabular}} & {\begin{tabular}[c]{@{}c@{}}MambaTS \\ (2024)\end{tabular}} & {\begin{tabular}[c]{@{}c@{}}iTransformer \\ (2024)\end{tabular}} & {\begin{tabular}[c]{@{}c@{}}FourierGNN \\ (2024)\end{tabular}} & {\begin{tabular}[c]{@{}c@{}}PatchTST \\ (2023)\end{tabular}} &  {\begin{tabular}[c]{@{}c@{}}Dlinear \\ (2023)\end{tabular}} & {\begin{tabular}[c]{@{}c@{}}S4 \\ (2022)\end{tabular}} \\ 
			\midrule
\multicolumn{9}{c}{{VitalDB $\to$ MOVER-SIS}} \\ 
			\midrule
\multicolumn{1}{c|}{\multirow{4}{*}{\rotatebox{90}{AVG}}} & \multicolumn{1}{c|}{MAE} & \cellcolor[RGB]{255,235,245} \textbf{0.048$\pm$0.005} &                     \cellcolor{blue!8} 0.051$\pm$0.005 &                     0.063$\pm$0.003 &  0.070$\pm$0.003 &  0.064$\pm$0.009 &   0.077$\pm$0.004 &              0.067$\pm$0.009 \\ 
			\multicolumn{1}{c|}{} & \multicolumn{1}{c|}{RMSE} & \cellcolor[RGB]{255,235,245} \textbf{0.097$\pm$0.005} &                     \cellcolor{blue!8} 0.103$\pm$0.005 &                     0.116$\pm$0.003 &  0.123$\pm$0.004 &  0.135$\pm$0.009 &   0.129$\pm$0.005 &              0.138$\pm$0.010 \\ 
			\multicolumn{1}{c|}{} & \multicolumn{1}{c|}{R2} & \cellcolor[RGB]{255,235,245} \textbf{0.835$\pm$0.254} &                     \cellcolor{blue!8} 0.812$\pm$0.212 &                     0.771$\pm$0.191 &  0.745$\pm$0.183 &  0.713$\pm$0.321 &   0.740$\pm$0.256 &              0.705$\pm$0.303 \\ 
			\multicolumn{1}{c|}{} & \multicolumn{1}{c|}{CC} & \cellcolor[RGB]{255,235,245} \textbf{0.832$\pm$0.049} &                     \cellcolor{blue!8} 0.816$\pm$0.047 &                     0.783$\pm$0.041 &  0.752$\pm$0.033 &  0.777$\pm$0.037 &   0.779$\pm$0.035 &  0.767$\pm$0.033 \\ 
			
\midrule
			
\multicolumn{9}{c}{{MOVER-SIS $\to$ VitalDB}} \\ 
			\midrule
			
			\multicolumn{1}{c|}{\multirow{4}{*}{\rotatebox{90}{AVG}}} & \multicolumn{1}{c|}{MAE} & \cellcolor[RGB]{255,235,245} \textbf{0.046$\pm$0.004} &                     \cellcolor[RGB]{255,235,245} \textbf{0.046$\pm$0.004} &                     \cellcolor{blue!8} 0.051$\pm$0.004 &  0.054$\pm$0.003 &  0.058$\pm$0.005 &   0.067$\pm$0.005 &              0.063$\pm$0.005 \\ 
			\multicolumn{1}{c|}{} & \multicolumn{1}{c|}{RMSE} & \cellcolor[RGB]{255,235,245} \textbf{0.075$\pm$0.004} &                     \cellcolor{blue!8} 0.077$\pm$0.005 &                     0.083$\pm$0.004 &  0.086$\pm$0.004 &  0.108$\pm$0.007 &   0.114$\pm$0.006 &              0.112$\pm$0.007 \\ 
			\multicolumn{1}{c|}{} & \multicolumn{1}{c|}{R2} & \cellcolor[RGB]{255,235,245} \textbf{0.723$\pm$0.581} &                     \cellcolor{blue!8} 0.700$\pm$0.457 &                     0.549$\pm$0.837 &  0.573$\pm$1.197 &  0.354$\pm$0.644 &   0.198$\pm$1.817 &              0.239$\pm$0.827 \\ 
			\multicolumn{1}{c|}{} & \multicolumn{1}{c|}{CC} & \cellcolor[RGB]{255,235,245} \textbf{0.809$\pm$0.060} &                     \cellcolor{blue!8} 0.808$\pm$0.062 &                     0.790$\pm$0.058 &  0.772$\pm$0.054 &  0.775$\pm$0.048 &   0.763$\pm$0.049 &  0.763$\pm$0.045 \\ 
			
			\bottomrule
			
\end{tabular}
	\end{adjustbox}
\end{table*} 
\noindent \textbf{Implementation Details.} Our benchmark was developed primarily using the Time-Series-Library framework~\cite{wu2023timesnet}. We employed grid search to determine the optimal hyperparameters for each model. For training, we used the Adam optimizer~\cite{kingma2014adam} with a total of 10 epochs and a patience of 3 epochs. This configuration was found to be sufficient for model convergence.

\subsection{Benchmark Results}

\noindent \textbf{Track 1: Complete Variables Data.} Table~\ref{tab:main_track} reports the average model performance across four prediction horizons, using MAE, RMSE,  $R^2$, and CC as evaluation metrics. 
Errors are expressed in the original units of each variable (e.g., beats per minute for heart rate, mmHg for blood pressure, \% for SpO$_2$), ensuring clinical interpretability. See Tables~\ref{tab:variables_vitaldb}--\ref{tab:variables_mover} for more variable definitions and units.

On the VitalDB dataset, MambaTS consistently achieved the best overall performance across all metrics, followed by iTransformer and LightTS. In contrast, model performance on MOVER-SIS showed more variability across tasks and metrics, reflecting higher data heterogeneity. Nonetheless, MambaTS remained a strong performer, with FourierGNN and LightTS also showing competitive results.

Interestingly, variable-independent models (S4, DLinear, PatchTST) consistently underperformed, highlighting the importance of modeling inter-variable dependencies in multivariate vital sign forecasting. Physiological signals are often tightly coupled due to underlying regulatory mechanisms (e.g., cardio-respiratory interactions), and ignoring these relationships can hinder a model's capacity to capture the temporal patient dynamics. Therefore, our findings indicate that models incorporating variable dependencies are better suited for vital sign prediction tasks and tend to achieve superior performance.

\noindent \textbf{Track 2: Incomplete Variables Data.} This track places more stringent demands on models by requiring them to handle variable numbers of input features across samples. Consequently, fixed-dimension methods (LightTS and MICN) cannot be evaluated in this track. Their exclusion is not intended to bias results, but rather reflects a realistic challenge: intraoperative data often contains missing or intermittently available sensor signals, which these models cannot accommodate.

As a result of increased data incompleteness and heterogeneity, Table~\ref{tab:main_masked_track} shows a modest performance decline compared to Track 1 (Table~\ref{tab:main_track}). Despite these challenges, MambaTS consistently outperforms other models, followed by iTransformer and FourierGNN, reflecting their superior capacity to adapt to partial and irregular inputs. Conversely, variable-independent models like S4, DLinear, and PatchTST experience substantial performance drops, highlighting the necessity of modeling inter-variable dependencies to capture physiological couplings under missingness.

\begin{figure*}[!t]
	\centering
	\includegraphics[width=0.78\linewidth]{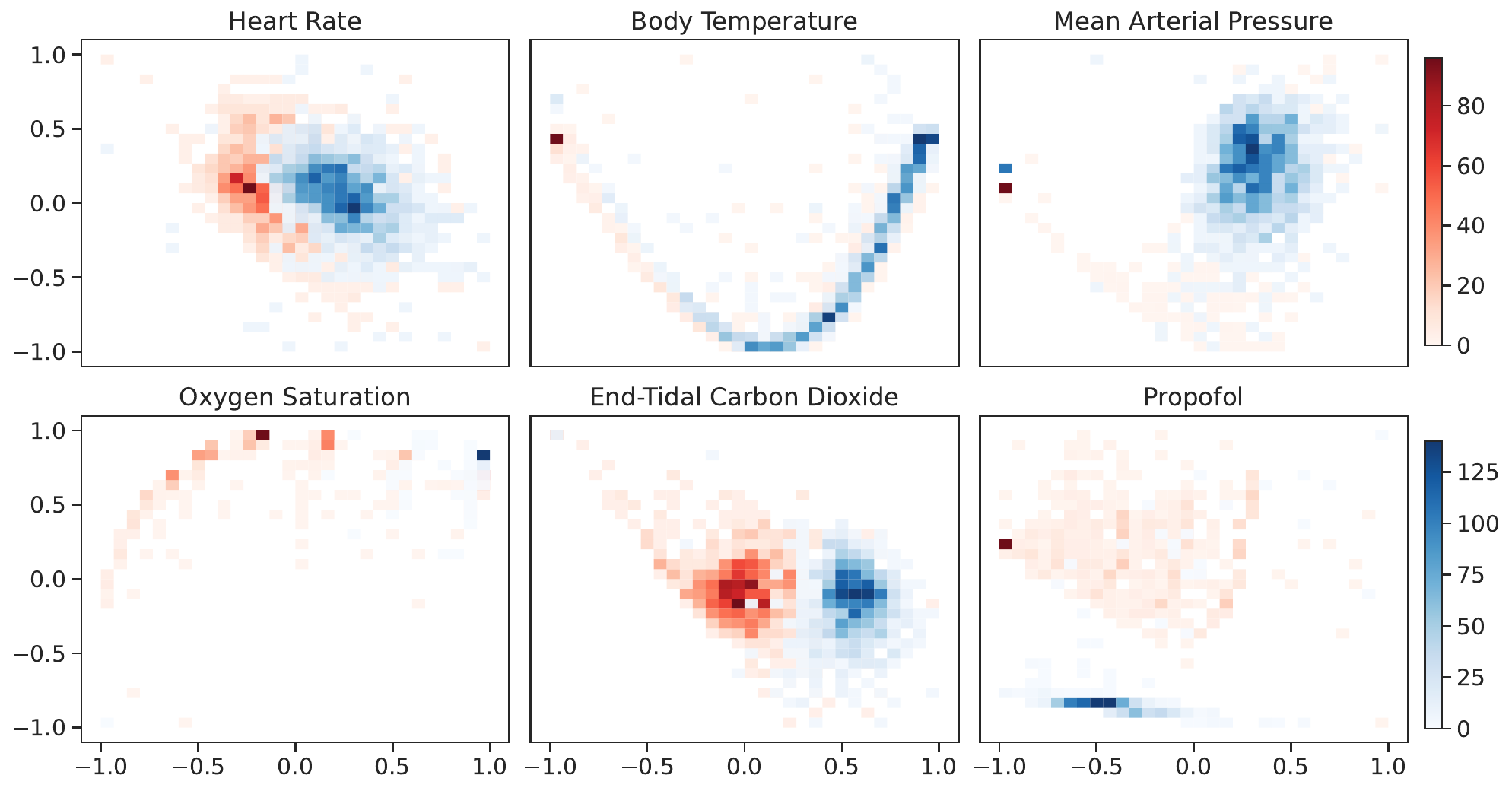}
	\caption{PCA visualization of representative dynamic variables across datasets. Each subplot shows the first and second principal components for a selected variable, illustrating distributional differences between VitalDB (blue) and MOVER-SIS (red).}
	\label{fig:pca}
\end{figure*}

\noindent \textbf{Track 3: Cross-Center Generalization.} The robustness of vital sign prediction models is essential for ensuring their applicability across diverse clinical settings. As illustrated in Figure~\ref{fig:pca}, principal component analysis (PCA)~\cite{wold1987principal} reveals substantial differences in the distribution of variables between the VitalDB and MOVER-SIS datasets, highlighting the challenges posed by cross-center validation.

\begin{figure*}[!t]
	\centering
	\includegraphics[width=0.85\linewidth]{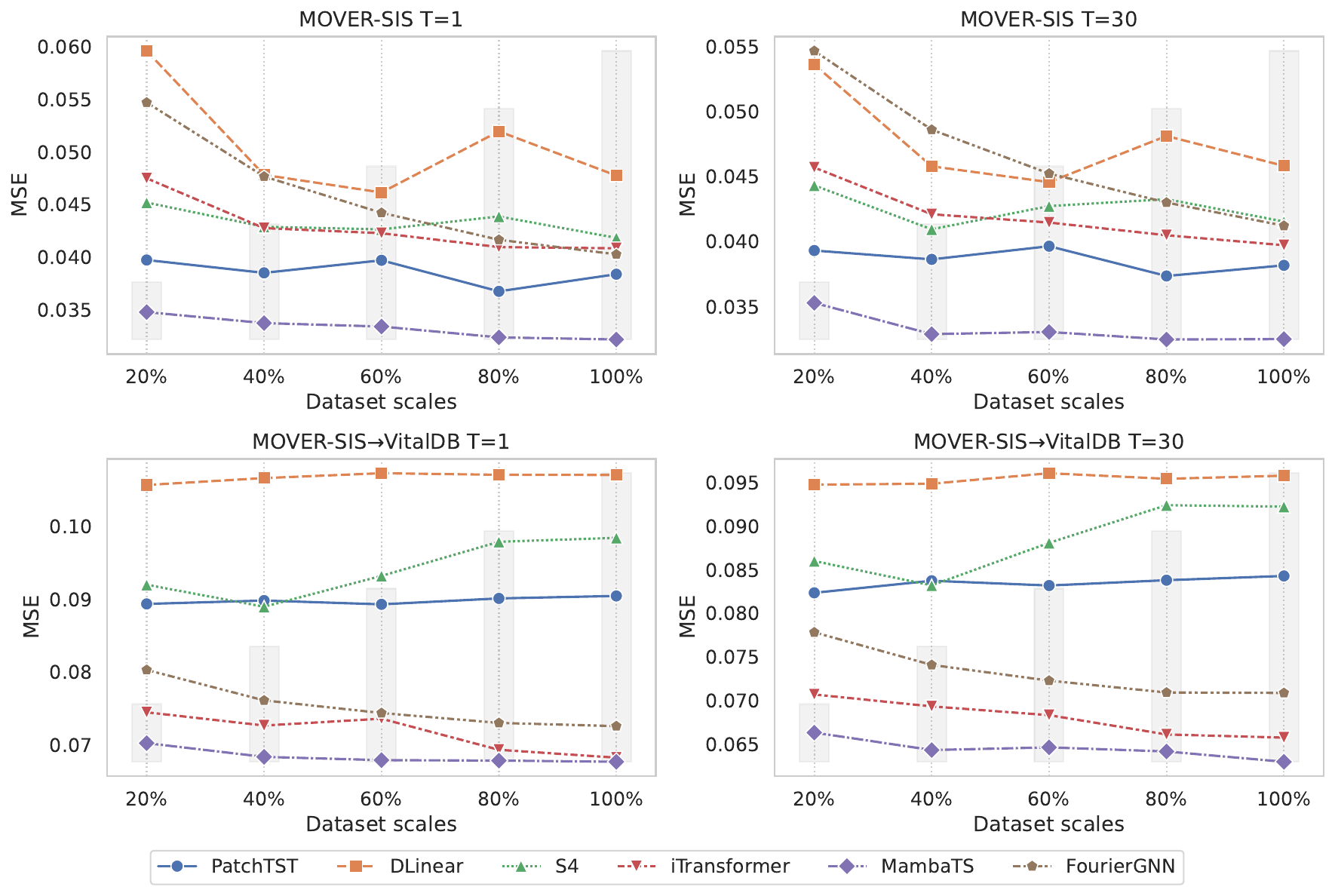}
	\caption{Performance comparison of different models on datasets of varying scales.}
	\label{fig:data_scale}
\end{figure*}

\begin{figure*}[!t]
	\centering
	\includegraphics[width=0.93\linewidth]{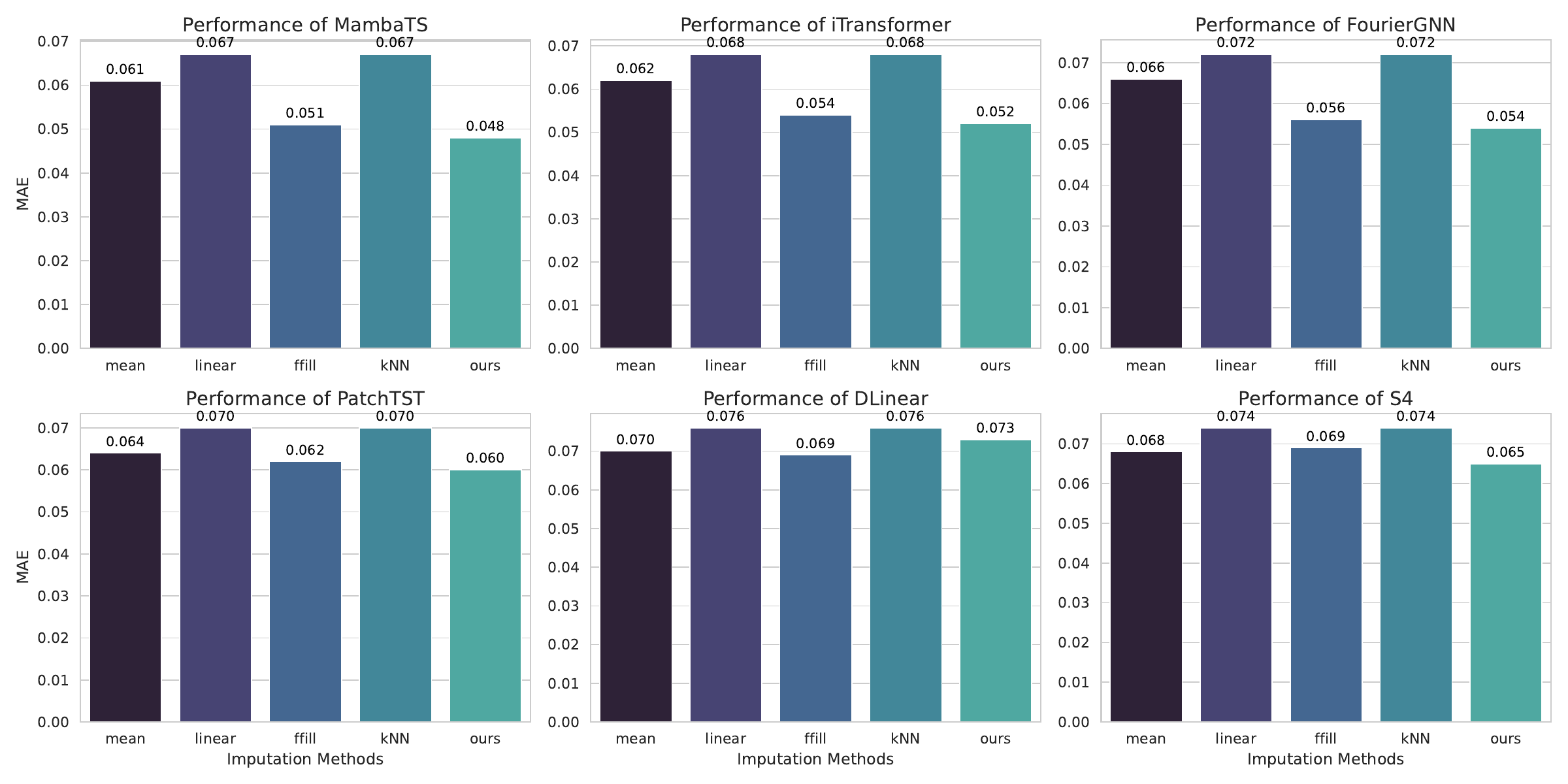}
	\caption{Comparison of model performance under different data imputation strategies.}
	\label{fig:impute}
\end{figure*}

To evaluate cross-center generalization, we conducted two transfer experiments: training on VitalDB and testing on MOVER-SIS, and vice versa. For reference, we also report an upper-bound performance, obtained by training MambaTS directly on the target domain, given its strong performance in both Track 1 and Track 2. As shown in Table~\ref{tab:main_dg_track}, MambaTS consistently outperformed other baseline models in both directions, with iTransformer performing competitively. Notably, in the MOVER-SIS $\to$ VitalDB setting, MambaTS achieved performance close to that obtained when trained and tested directly on VitalDB, indicating strong generalization with minimal performance degradation. 

In contrast, the VitalDB $\to$ MOVER-SIS transfer exhibited a larger performance gap, which is likely attributable to the smaller training sample size of VitalDB (962 surgeries) compared to MOVER-SIS (3,221 surgeries). This discrepancy highlights the critical importance of data scale and diversity in learning generalizable representations. These findings suggest that models trained on sufficiently large and heterogeneous cohorts may effectively transfer to new clinical environments with minimal adaptation. Therefore, expanding training datasets through multi-center collaborations holds strong potential to improve both accuracy and robustness of clinical time-series prediction models in real-world deployments.

\subsection{Ablation Studies}
\label{sec:ablation}

\noindent \textbf{Impact of Data Scale.} Extending Track~3 findings, we examined how training size affects model performance (Figure~\ref{fig:data_scale}). Variable-dependent models (iTransformer, FourierGNN, MambaTS) consistently improve with larger datasets, reflecting their ability to exploit inter-variable relationships essential for robustness across diverse clinical contexts. In contrast, variable-independent models (DLinear, PatchTST, S4) decline as data scale grows, highlighting the limitations of treating each variable independently. These results underscore the critical role of modeling inter-variable dynamics for reliable performance on large, heterogeneous clinical data.

\noindent \textbf{Impact of Imputation.} In this section, we analyze the effect of data imputation in VitalBench. As we previously argued, data missingness is an inherent challenge in clinical settings (e.g., prolonged sensor dropouts, as shown in Figure~\ref{fig:missing_rate}). In such cases, imputation can introduce spurious values, distorting predictions of vital signs. To address this, we employ a masked loss function that naturally handles missingness without altering the original data, preserving clinical validity.

Figure~\ref{fig:impute} compares model performance across four common imputation methods: mean imputation, linear interpolation, forward-fill (ffill), and k-nearest neighbors (kNN). Across nearly all cases, masked loss outperforms these strategies, achieving superior predictive accuracy. DLinear is an exception, showing sensitivity to imputed zeros. While ffill performs reasonably, its reliance on previous values is only valid over short gaps, whereas clinical sensor dropouts often persist longer. These results highlight the importance of explicitly handling missingness rather than relying on imputation for robust and clinically meaningful predictions.

\noindent \textbf{Discussion.}
Building on our findings, future research in intraoperative vital sign prediction should prioritize models that capture complex inter-variable dependencies, which are critical for both predictive accuracy and robustness. These models must also accommodate variable input dimensionality to reflect patient heterogeneity and ensure strong cross-center generalization. Equally important is expanding data diversity and scale, as larger multi-center datasets not only enhance performance but also improve generalizability across clinical settings and populations. Integrating these strategies will be essential to develop robust, clinically reliable prediction systems with broad applicability in real-world healthcare.
 \section{Conclusion}

In this study, we present VitalBench, a comprehensive benchmark designed to evaluate intraoperative vital sign forecasting models under clinically realistic conditions. By addressing challenges such as incomplete data and cross-center variability, VitalBench provides a rigorous framework for developing robust and generalizable predictive models.

Our experiments yield three key insights. First, accurately modeling inter-variable dependencies is critical. Models such as MambaTS, which capture physiological interactions among vital signs, consistently outperform variable-independent baselines. Second, using a masked loss function that naturally accommodates missing values leads to more reliable predictions than traditional imputation methods, which often introduce artificial noise. Third, increasing the scale and diversity of training data improves generalization performance, but substantial performance gaps remain across clinical centers. This highlights the complexity introduced by variations in patient populations, monitoring protocols, and institutional practices.

VitalBench thus serves as a step toward bridging the gap between algorithmic innovation and clinical deployment. Future efforts should focus on developing models that adapt to varying input dimensions, explicitly handle uncertainty due to missingness, and leverage multi-center datasets to enhance robustness across real-world healthcare environments.
 
\bibliographystyle{IEEEtran}
\IEEEtriggeratref{48}
\bibliography{main}{}

\clearpage

\appendices

\noindent
\begin{center}
	\Large \textbf{Supplementary Materials for VitalBench: A Rigorous Multi-Center Benchmark for Long-Term Vital Sign Prediction\\in Intraoperative Care}
\end{center}

\section*{VitalDB Dataset}

The \textbf{VitalDB dataset}~\cite{lee2022vitaldb} is a publicly available, single-center perioperative dataset that includes detailed anesthesia records from 6,388 non-cardiac surgeries performed between August 2016 and June 2017 at Seoul National University Hospital, South Korea. The dataset provides comprehensive perioperative monitoring data, including physiological signals, anesthetic parameters, and demographic information. VitalDB is accessible at \href{https://vitaldb.net/}{\texttt{https://vitaldb.net/}}, offering open access to synchronized intraoperative waveforms, trends, and event data for research and educational purposes.

\subsection*{Inclusion and Exclusion Criteria}

The following criteria were applied for patient inclusion:

\begin{itemize}
	\item Surgery duration of at least 2 hours
	\item General anesthesia administered throughout the procedure
	\item Age of 18 years or older
	\item Weight greater than 35 kg
\end{itemize}

After applying these criteria, 962 surgeries were included for analysis. The statistical information regarding the dataset division is presented in Table~\ref{tab:summary_cohorts_vitaldb}.

\subsection*{Variables in the Dataset}

As shown in Table~\ref{tab:variables_vitaldb}, each surgery record in the VitalDB dataset includes 17 variables, divided into the following categories:

\begin{itemize}
	\item \textbf{Demographic Data}: 'Age', 'Gender', 'Weight', 'Height', 'ASA'
	\item \textbf{Anesthetic Information}: 'Orchestra/PPF20\_VOL', 'Orchestra/RFTN20\_VOL', 'Orchestra/PPF20\_CE', 'Orchestra/\\RFTN20\_CE'
	\item \textbf{Vital Signs}: 'Solar8000/ART\_DBP', 'Solar8000/ART\_SBP', 'Solar8000/ART\_MBP', 'Solar8000/HR', 'Solar8000/BT', 'Solar8000/ETCO2', 'Solar8000/PLETH\_SPO2', 'BIS/BIS'
\end{itemize}

\begin{figure*}[!t]
	\centering
	\includegraphics[width=1\linewidth]{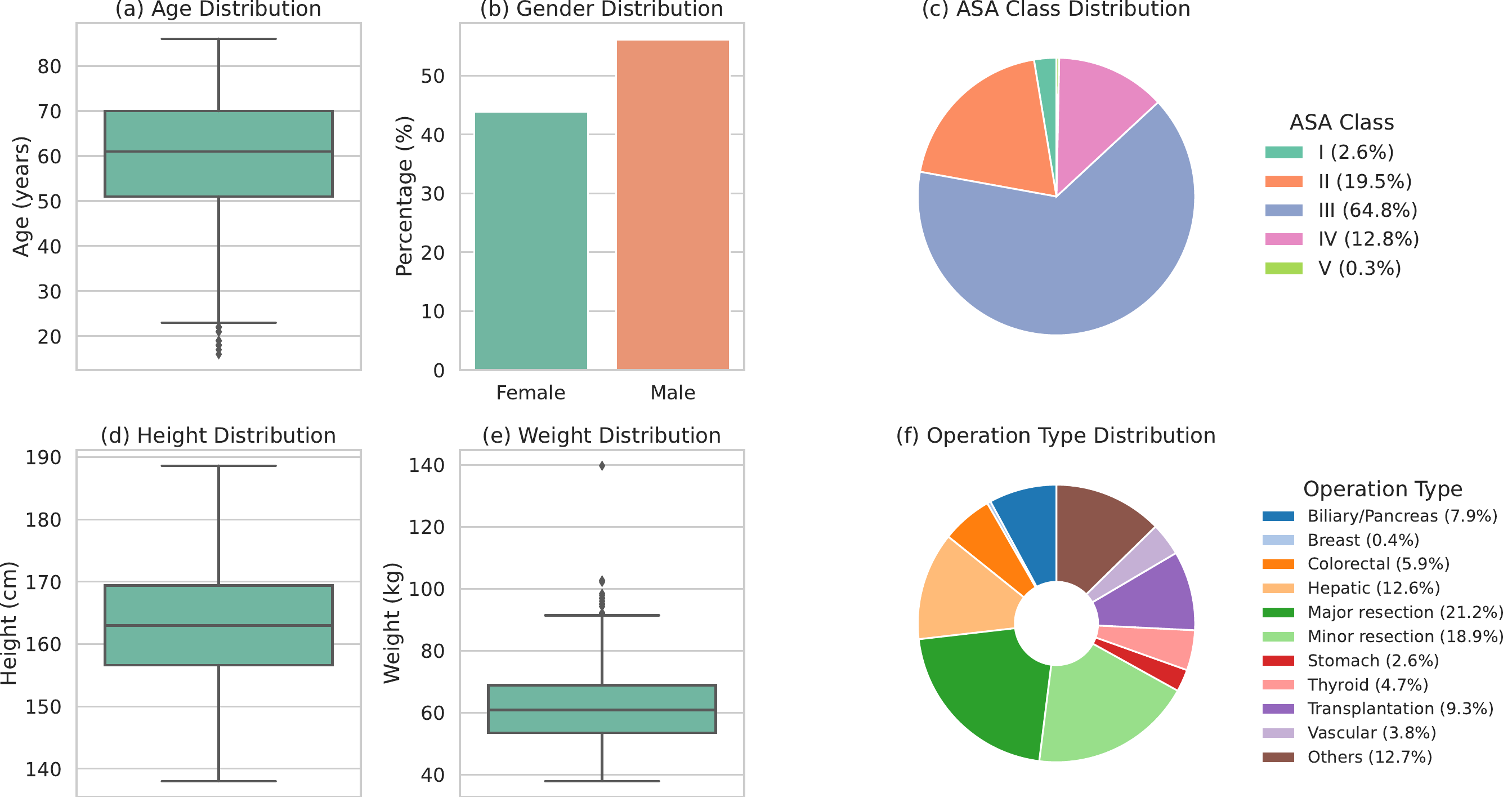}
	\caption{Summary of patient demographics and surgical characteristics in the VitalDB cohort. Box plots illustrate the distributions of age (a), height (d) and weight (e), while bar plots summarize the gender (b) distributions. The pie charts display the proportions of ASA classes (c) and operation types (f).}
	
	\label{fig:vital_db}
\end{figure*}

Among these, demographic variables serve as constant features and act as covariates for modeling other dynamic variables, with operation type also reported for reference. The overall demographic and surgical characteristics of the VitalDB cohort are summarized in Figure~\ref{fig:vital_db}.

\newpage

\section*{MOVER-SIS Dataset}

The \textbf{MOVER-SIS dataset}, a subset of the largest publicly available perioperative dataset~\cite{samad2023medical}, includes data from 19,114 patients who underwent surgeries at the University of California, Irvine Medical Center between 2015 and 2017. It contains rich perioperative information, such as demographic details, medication records, ventilator parameters, and vital signs collected through the Surgical Information System (SIS). The dataset is available at \href{https://mover.ics.uci.edu/}{\texttt{https://mover.ics.uci.edu/}} and was used under the UCI-OR Data Use Agreement.

\subsection*{Inclusion and Exclusion Criteria}

The inclusion and exclusion criteria for the MOVER-SIS dataset are identical to those used in the VitalDB dataset:

\begin{itemize}
	\item Surgery duration of at least 2 hours
	\item General anesthesia administered throughout the procedure
	\item Age of 18 years or older
	\item Weight greater than 35 kg
\end{itemize}

The final dataset contains records from all eligible surgeries meeting these criteria. The statistical information regarding the dataset division is presented in Table~\ref{tab:summary_cohorts_mover}.

\subsection*{Variables in the Dataset}

As shown in~\ref{tab:variables_mover}, each surgery record in the MOVER-SIS dataset includes 33 variables, categorized as follows:

\begin{itemize}
	\item \textbf{Demographic Data}: 'Age', 'Gender', 'Height', 'Weight'
	\item \textbf{Anesthetic Information}: 'Agent\_D', 'Agent\_I', 'Agent\_N', 'Agent\_S', 'Agent\_Fi', 'Agent\_Et', 'Ephedrine', 'Fentanyl', 'Hydromorphone', 'Lidocaine', 'Midazolam', 'Ondansetron', 'Phenylephrine', 'Plasmalyte', 'Propofol', 'Vecuronium'
	\item \textbf{Vital Signs}: 'HRp', 'SP02', 'Temp', 'SBP\_ART', 'MAP\_ART', 'DBP\_ART'
	\item \textbf{Ventilator Parameters}: 'TV', 'RR', 'PEEP', 'PIP', 'ETCO2', 'FI02', 'ET02'
\end{itemize}

\begin{figure*}[!t]
	\centering
	\includegraphics[width=1\linewidth]{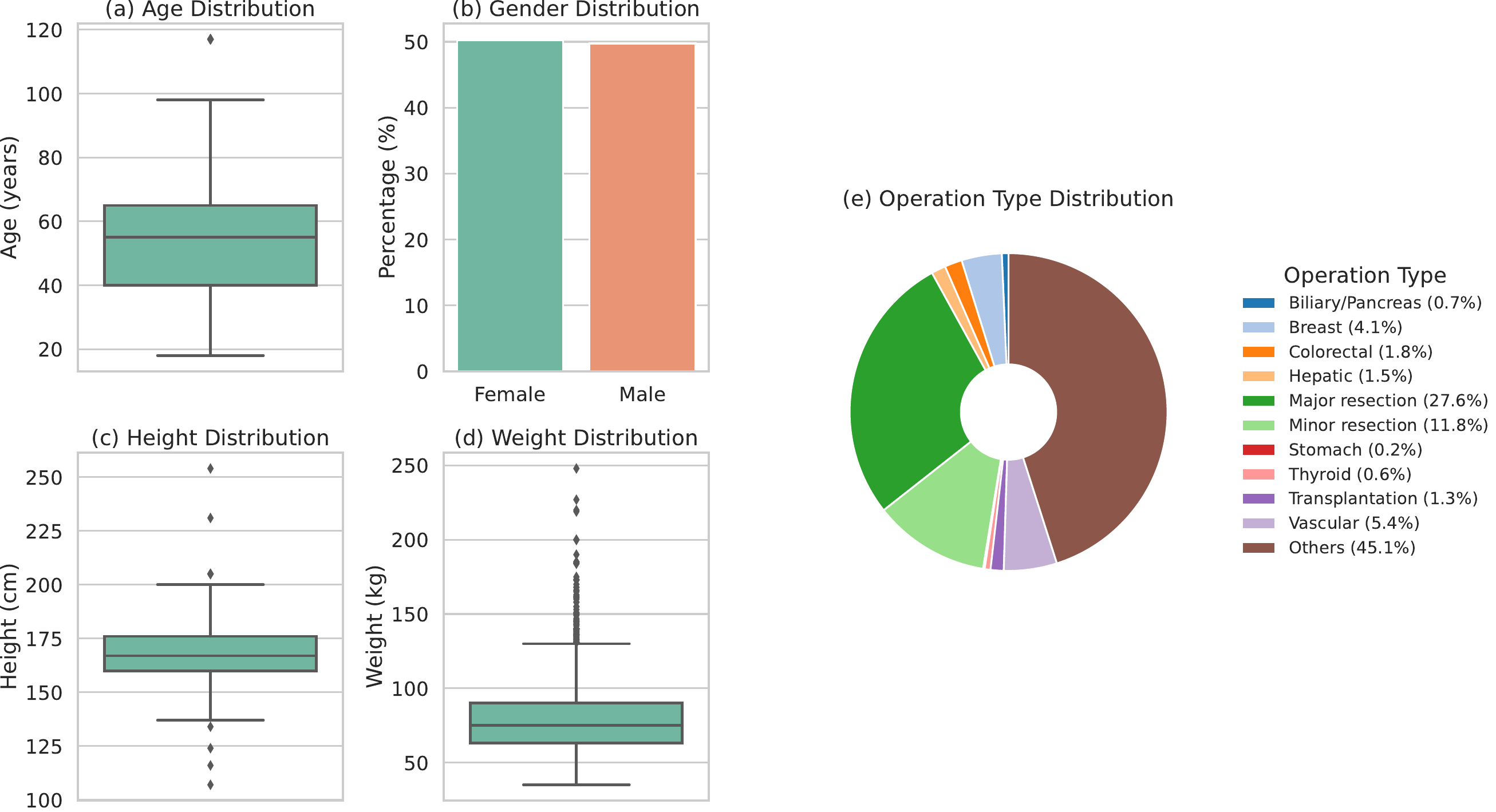}
	\caption{Summary of patient demographics and surgical characteristics in the MOVER-SIS cohort. Box plots illustrate the distributions of age (a), height (c) and weight (d), while bar plots summarize the gender (b) distributions. The pie chart presents the proportions of different operation types (e).}
	
	\label{fig:mover_sis}
\end{figure*}

Similar to the VitalDB cohort, demographic variables in the MOVER-SIS dataset remain constant throughout each surgery and serve as covariates for predicting other physiological and pharmacological signals, with operation type also reported for reference. Figure~\ref{fig:mover_sis} summarizes the demographic distributions and surgical characteristics of the MOVER-SIS cohort.

\begin{table*}[!t]
	\caption{Summary of Demographic and Clinical Variables across Different Cohorts in VitalDB.}
	\label{tab:summary_cohorts_vitaldb}
	\centering
	\begin{adjustbox}{max width=\linewidth}
		\begin{tabular}{@{}l c c c c c c@{}}
			\toprule
			\textbf{Variable} & \textbf{Overall} & \textbf{Training} & \textbf{Validation} & \textbf{Test} & \textbf{P-Value} & \textbf{P-Test} \\ \midrule
			\textbf{n} & 962 & 673 & 96 & 193 &  &  \\ \midrule
			\textbf{Age, median [Q1, Q3]} & 61.0 [51.0, 70.0] & 61.0 [52.0, 70.0] & 60.5 [53.8, 70.0] & 60.0 [51.0, 70.0] & 0.962 & Kruskal-Wallis \\
			\textbf{Gender, n (\%)} &  &  &  &  &  &  \\
			\hspace{1em} Female & 422 (43.9) & 296 (44.0) & 48 (50.0) & 78 (40.4) & 0.301 & Chi-squared \\
			\hspace{1em} Male & 540 (56.1) & 377 (56.0) & 48 (50.0) & 115 (59.6) &  &  \\ 
\textbf{Height, median [Q1, Q3]} & 163.0 [156.6, 169.4] & 163.0 [156.5, 169.3] & 161.2 [154.4, 169.9] & 164.1 [157.3, 170.0] & 0.175 & Kruskal-Wallis \\
			\textbf{Weight, median [Q1, Q3]} & 60.8 [53.5, 68.9] & 60.8 [53.8, 68.2] & 58.6 [51.1, 70.3] & 62.2 [54.0, 71.2] & 0.272 & Kruskal-Wallis \\
			\textbf{ASA, n (\%)} &  &  &  &  &  &  \\
			\hspace{1em} I & 25 (2.6) & 15 (2.2) & 2 (2.1) & 8 (4.1) & 0.412 & Chi-squared \\
			\hspace{1em} II & 188 (19.5) & 131 (19.5) & 21 (21.9) & 36 (18.7) &  &  \\
			\hspace{1em} III & 623 (64.8) & 445 (66.1) & 55 (57.3) & 123 (63.7) &  &  \\
			\hspace{1em} IV & 123 (12.8) & 81 (12.0) & 17 (17.7) & 25 (13.0) &  &  \\
			\hspace{1em} V & 3 (0.3) & 1 (0.1) & 1 (1.0) & 1 (0.5) &  &  \\
			
			\textbf{Operation Type, n (\%)} &  &  &  &  &  &  \\
			\hspace{1em} Biliary/Pancreas & 76 (7.9) & 46 (6.8) & 11 (11.5) & 19 (9.8) & 0.554 & Chi-squared \\
			\hspace{1em} Breast & 4 (0.4) & 3 (0.4) & 1 (1.0) &  &  &  \\
			\hspace{1em} Colorectal & 57 (5.9) & 39 (5.8) & 7 (7.3) & 11 (5.7) &  &  \\
			\hspace{1em} Hepatic & 121 (12.6) & 80 (11.9) & 14 (14.6) & 27 (14.0) &  &  \\
			\hspace{1em} Major resection & 204 (21.2) & 140 (20.8) & 25 (26.0) & 39 (20.2) &  &  \\
			\hspace{1em} Minor resection & 182 (18.9) & 138 (20.5) & 12 (12.5) & 32 (16.6) &  &  \\			
			\hspace{1em} Stomach & 25 (2.6) & 16 (2.4) & 4 (4.2) & 5 (2.6) &  &  \\
			\hspace{1em} Thyroid & 45 (4.7) & 30 (4.5) & 5 (5.2) & 10 (5.2) &  &  \\
			\hspace{1em} Transplantation & 89 (9.3) & 61 (9.1) & 6 (6.2) & 22 (11.4) &  &  \\
			\hspace{1em} Vascular & 37 (3.8) & 25 (3.7) & 4 (4.2) & 8 (4.1) &  &  \\
			\hspace{1em} Others & 122 (12.7) & 95 (14.1) & 7 (7.3) & 20 (10.4) &  &  \\
			
			\bottomrule
		\end{tabular}
	\end{adjustbox}
\end{table*}

\begin{table*}[!ht]
	\caption{Summary of Demographic and Clinical Variables across Different Cohorts in MOVER-SIS.}
	\label{tab:summary_cohorts_mover}
	\centering
	\begin{adjustbox}{max width=\linewidth}
		\begin{tabular}{@{}l c c c c c c@{}}
			\toprule
			\textbf{Variable} & \textbf{Overall} & \textbf{Training} & \textbf{Validation} & \textbf{Test} & \textbf{P-Value} & \textbf{P-Test} \\ \midrule
			\textbf{n} & 3221 & 2254 & 322 & 645 &  &  \\ \midrule
			\textbf{Age, median [Q1, Q3]} & 55.0 [40.0, 65.0] & 54.0 [40.0, 65.0] & 57.0 [44.0, 66.0] & 55.0 [42.0, 66.0] & 0.235 & Kruskal-Wallis \\
			\textbf{Gender, n (\%)} &  &  &  &  &  &  \\ 
			\hspace{1em} Female & 1619 (50.3) & 1130 (50.1) & 159 (49.4) & 330 (51.2) & 0.850 & Chi-squared \\
			\hspace{1em} Male & 1602 (49.7) & 1124 (49.9) & 163 (50.6) & 315 (48.8) &  &  \\ 
\textbf{Height, median [Q1, Q3]} & 167.0 [160.0, 176.0] & 167.0 [160.0, 175.0] & 167.0 [160.0, 177.0] & 167.0 [160.0, 177.0] & 0.567 & Kruskal-Wallis \\
			\textbf{Weight, median [Q1, Q3]} & 75.0 [63.0, 90.0] & 75.0 [63.0, 90.0] & 76.5 [61.0, 92.8] & 75.0 [65.0, 88.0] & 0.854 & Kruskal-Wallis \\
			
			\textbf{Operation Type, n (\%)} &  &  &  &  &  &  \\
			\hspace{1em} Biliary/Pancreas & 22 (0.7) & 13 (0.6) & 4 (1.2) & 5 (0.8) & 0.965 & Chi-squared \\
			\hspace{1em} Breast & 132 (4.1) & 93 (4.1) & 13 (4.0) & 26 (4.0) &  &  \\
			\hspace{1em} Colorectal & 57 (1.8) & 42 (1.9) & 3 (0.9) & 12 (1.9) &  &  \\
			\hspace{1em} Hepatic & 47 (1.5) & 38 (1.7) & 5 (1.6) & 4 (0.6) &  &  \\
			\hspace{1em} Major resection & 889 (27.6) & 622 (27.6) & 90 (28.0) & 177 (27.4) &  &  \\
			\hspace{1em} Minor resection & 381 (11.8) & 262 (11.6) & 41 (12.7) & 78 (12.1) &  &  \\			
			\hspace{1em} Stomach & 5 (0.2) & 4 (0.2) &  & 1 (0.2) &  &  \\
			\hspace{1em} Thyroid & 19 (0.6) & 13 (0.6) & 3 (0.9) & 3 (0.5) &  &  \\
			\hspace{1em} Transplantation & 43 (1.3) & 28 (1.2) & 5 (1.6) & 10 (1.6) &  &  \\
			\hspace{1em} Vascular & 174 (5.4) & 118 (5.2) & 18 (5.6) & 38 (5.9) &  &  \\
			\hspace{1em} Others & 1452 (45.1) & 1021 (45.3) & 140 (43.5) & 291 (45.1) &  &  \\
			
			\bottomrule
		\end{tabular}
	\end{adjustbox}
\end{table*}

\begin{table*}[!t]
	\caption{Summary of Variables Selected from the VitalDB Dataset.}
	\label{tab:variables_vitaldb}
	\centering
	\begin{adjustbox}{max width=1\linewidth}
		\begin{tabular}{@{}l l l c@{}}
			\toprule
			\textbf{Variable} & \textbf{Description} & \textbf{Unit} & \textbf{Covariate} \\ \midrule
			Age                      & Age of the patient                        & years       & $\checkmark$ \\
			Gender                      & Gender of the patient                     & --          & $\checkmark$ \\
			Weight                   & Body weight of the patient                & kg          & $\checkmark$ \\
			Height                   & Height of the patient                     & cm          & $\checkmark$ \\
			ASA                      & ASA classification                        & -- & $\checkmark$ \\
			Orchestra/PPF20\_VOL     & Propofol infusion volume                  & mL          &               \\
			Orchestra/RFTN20\_VOL    & Remifentanil infusion volume              & mL          &               \\
			Orchestra/PPF20\_CE      & Propofol effect-site concentration        & $\mu$g/mL   &               \\
			Orchestra/RFTN20\_CE     & Remifentanil effect-site concentration    & ng/mL       &               \\
			Solar8000/HR             & Heart rate                                & bpm         &               \\
			Solar8000/BT             & Body temperature                          & °C          &               \\
			Solar8000/ART\_DBP       & Arterial diastolic blood pressure         & mmHg        &               \\
			Solar8000/ART\_SBP       & Arterial systolic blood pressure          & mmHg        &               \\
			Solar8000/ART\_MBP       & Arterial mean blood pressure              & mmHg        &               \\
			Solar8000/ETCO2          & End-tidal carbon dioxide                  & mmHg        &               \\
			Solar8000/PLETH\_SPO2    & Peripheral oxygen saturation              & \%          &               \\
			BIS/BIS                  & Bispectral index                          & [0--100]    &               \\
			\bottomrule
		\end{tabular}
	\end{adjustbox}
\end{table*}

\begin{table*}[h]
	\caption{Summary of Variables Selected from the MOVER-SIS Dataset.}
	\label{tab:variables_mover}
	\centering
	\begin{adjustbox}{max width=0.75\linewidth}
		\begin{tabular}{@{}l l l c@{}}
			\toprule
			\textbf{Variable} & \textbf{Description} & \textbf{Unit} & \textbf{Covariate} \\ \midrule
			Age           & Age of the patient                       & years       & $\checkmark$ \\
			Gender        & Gender of the patient                    & --          & $\checkmark$ \\
			Wt            & Body weight of the patient               & kg          & $\checkmark$ \\
			Ht            & Height of the patient                    & cm          & $\checkmark$ \\
			ASA           & ASA classification                      & -- & $\checkmark$ \\
			Agent\_D      & Dose of agent D                         & mL          &              \\
			Agent\_I      & Dose of agent I                         & mL          &              \\
			Agent\_N      & Dose of agent N                         & mL          &              \\
			Agent\_S      & Dose of agent S                         & mL          &              \\
			Agent\_Fi     & Inspired concentration of agent         & \%          &              \\
			Agent\_Et     & End-tidal concentration of agent        & \%          &              \\
			Ephedrine     & Ephedrine dose                          & mg          &              \\
			Fentanyl      & Fentanyl dose                           & $\mu$g      &              \\
			Hydromorphone & Hydromorphone dose                      & mg          &              \\
			Lidocaine     & Lidocaine dose                          & mg          &              \\
			Midazolam     & Midazolam dose                          & mg          &              \\
			Ondansetron   & Ondansetron dose                        & mg          &              \\
			Phenylephrine & Phenylephrine dose                      & $\mu$g      &              \\
			Plasmalyte    & Plasmalyte dose                         & mL          &              \\
			Propofol      & Propofol dose                           & mg          &              \\
			Vecuronium    & Vecuronium dose                         & mg          &              \\
			HRp           & Heart rate (postoperative)              & bpm         &              \\
			SpO2          & Peripheral oxygen saturation             & \%          &              \\
			Temp          & Body temperature                        & °C          &              \\
			SBP\_ART      & Systolic blood pressure                 & mmHg        &              \\
			MAP\_ART      & Mean arterial pressure                  & mmHg        &              \\
			DBP\_ART      & Diastolic blood pressure               & mmHg        &              \\
			TV            & Tidal volume                           & mL          &              \\
			RR            & Respiratory rate                        & breaths/min &              \\
			PEEP          & Positive end-expiratory pressure        & cmH\textsubscript{2}O &      \\
			PIP           & Peak inspiratory pressure               & cmH\textsubscript{2}O &      \\
			ETCO2         & End-tidal CO\textsubscript{2}          & mmHg        &              \\
			FI02          & Inspired oxygen concentration           & \%          &              \\
			ET02          & End-tidal oxygen concentration          & \%          &              \\
			\bottomrule
		\end{tabular}
	\end{adjustbox}
\end{table*}

\begin{table*}[!ht]
	\fontsize{10pt}{12pt}\selectfont
	\centering
	\renewcommand{\arraystretch}{1.1}
	\caption{Full performance comparison of different models on the complete variables track. The historical window is set to 30 minutes, and the forecasting horizons are 1, 3, 10, and 30 minutes.
	}
	\label{tab:main_track_full}
	\begin{adjustbox}{max width=\linewidth}
		\begin{tabular}{@{}cccccccccc@{}}
			\toprule
\multicolumn{2}{c}{{\diagbox[width=8em, height=3.9em]{Metrics}{Models}}}                                                   & {\begin{tabular}[c]{@{}c@{}}MambaTS\\ (2024)\end{tabular}} & {\begin{tabular}[c]{@{}c@{}}LightTS\\ (2023)\end{tabular}} & {\begin{tabular}[c]{@{}c@{}}iTransformer\\ (2024)\end{tabular}} & {\begin{tabular}[c]{@{}c@{}}FourierGNN\\ (2024)\end{tabular}} & {\begin{tabular}[c]{@{}c@{}}PatchTST\\ (2023)\end{tabular}} & {\begin{tabular}[c]{@{}c@{}}MICN\\ (2024)\end{tabular}} & {\begin{tabular}[c]{@{}c@{}}DLinear\\ (2023)\end{tabular}} & {\begin{tabular}[c]{@{}c@{}}S4\\ (2022)\end{tabular}} \\ 
			\midrule
\multicolumn{10}{c}{{VitalDB}} \\ 
			\midrule
\multicolumn{1}{c|}{\multirow{4}{*}{1}} & \multicolumn{1}{c|}{MAE} & \cellcolor[RGB]{255,235,245} \textbf{0.022$\pm$0.007} &                     \cellcolor{blue!8} 0.026$\pm$0.007 &                                        0.027$\pm$0.006 &                                        0.031$\pm$0.009 &  0.038$\pm$0.073 &   0.031$\pm$0.009 &                     0.028$\pm$0.007 &               0.034$\pm$0.008 \\ 
			\multicolumn{1}{c|}{} & \multicolumn{1}{c|}{RMSE} & \cellcolor[RGB]{255,235,245} \textbf{0.041$\pm$0.009} &                                        0.046$\pm$0.010 &                     \cellcolor{blue!8} 0.045$\pm$0.010 &                                        0.052$\pm$0.014 &  0.090$\pm$0.072 &   0.062$\pm$0.016 &                     0.061$\pm$0.013 &               0.063$\pm$0.012 \\ 
			\multicolumn{1}{c|}{} & \multicolumn{1}{c|}{R2} & \cellcolor[RGB]{255,235,245} \textbf{0.977$\pm$0.013} &                                        0.971$\pm$0.014 &                     \cellcolor{blue!8} 0.973$\pm$0.013 &                                        0.962$\pm$0.022 &  0.844$\pm$0.892 &   0.942$\pm$0.028 &                     0.946$\pm$0.024 &               0.941$\pm$0.024 \\ 
			\multicolumn{1}{c|}{} & \multicolumn{1}{c|}{CC} & \cellcolor[RGB]{255,235,245} \textbf{0.930$\pm$0.083} &                                        0.918$\pm$0.085 &                     \cellcolor{blue!8} 0.927$\pm$0.084 &                                        0.895$\pm$0.115 &  0.866$\pm$0.192 &   0.905$\pm$0.084 &                     0.914$\pm$0.088 &  0.899$\pm$0.092 \\ \cline{1-10} 
			\multicolumn{1}{c|}{\multirow{4}{*}{3}} & \multicolumn{1}{c|}{MAE} & \cellcolor[RGB]{255,235,245} \textbf{0.028$\pm$0.010} &                     \cellcolor{blue!8} 0.034$\pm$0.010 &                     \cellcolor{blue!8} 0.034$\pm$0.008 &                                        0.035$\pm$0.012 &  0.044$\pm$0.072 &   0.036$\pm$0.011 &                     0.040$\pm$0.010 &               0.040$\pm$0.010 \\ 
			\multicolumn{1}{c|}{} & \multicolumn{1}{c|}{RMSE} & \cellcolor[RGB]{255,235,245} \textbf{0.051$\pm$0.012} &                                        0.056$\pm$0.013 &                     \cellcolor{blue!8} 0.053$\pm$0.013 &                                        0.058$\pm$0.018 &  0.108$\pm$0.071 &   0.073$\pm$0.018 &                     0.077$\pm$0.015 &               0.075$\pm$0.015 \\ 
			\multicolumn{1}{c|}{} & \multicolumn{1}{c|}{R2} & \cellcolor[RGB]{255,235,245} \textbf{0.965$\pm$0.019} &                                        0.955$\pm$0.022 &                     \cellcolor{blue!8} 0.961$\pm$0.020 &                                        0.952$\pm$0.032 &  0.755$\pm$1.551 &   0.919$\pm$0.038 &                     0.914$\pm$0.034 &               0.916$\pm$0.037 \\ 
			\multicolumn{1}{c|}{} & \multicolumn{1}{c|}{CC} & \cellcolor[RGB]{255,235,245} \textbf{0.899$\pm$0.123} &                                        0.883$\pm$0.126 &                     \cellcolor{blue!8} 0.896$\pm$0.126 &                                        0.870$\pm$0.149 &  0.835$\pm$0.218 &   0.874$\pm$0.126 &                     0.875$\pm$0.131 &  0.863$\pm$0.134 \\ \cline{1-10} 
			\multicolumn{1}{c|}{\multirow{4}{*}{10}} & \multicolumn{1}{c|}{MAE} & \cellcolor[RGB]{255,235,245} \textbf{0.035$\pm$0.012} &                                        0.042$\pm$0.013 &                     \cellcolor{blue!8} 0.040$\pm$0.012 &                                        0.044$\pm$0.015 &  0.064$\pm$0.054 &   0.048$\pm$0.015 &                     0.061$\pm$0.013 &               0.054$\pm$0.013 \\ 
			\multicolumn{1}{c|}{} & \multicolumn{1}{c|}{RMSE} & \cellcolor[RGB]{255,235,245} \textbf{0.061$\pm$0.017} &                                        0.068$\pm$0.018 &                     \cellcolor{blue!8} 0.065$\pm$0.018 &                                        0.070$\pm$0.023 &  0.138$\pm$0.059 &   0.092$\pm$0.025 &                     0.102$\pm$0.019 &               0.097$\pm$0.022 \\ 
			\multicolumn{1}{c|}{} & \multicolumn{1}{c|}{R2} & \cellcolor[RGB]{255,235,245} \textbf{0.947$\pm$0.036} &                                        0.931$\pm$0.040 &                     \cellcolor{blue!8} 0.940$\pm$0.038 &                                        0.926$\pm$0.058 &  0.570$\pm$3.051 &   0.864$\pm$0.069 &                     0.844$\pm$0.060 &               0.851$\pm$0.069 \\ 
			\multicolumn{1}{c|}{} & \multicolumn{1}{c|}{CC} & \cellcolor[RGB]{255,235,245} \textbf{0.850$\pm$0.185} &                                        0.832$\pm$0.188 &                     \cellcolor{blue!8} 0.847$\pm$0.188 &                                        0.819$\pm$0.211 &  0.763$\pm$0.249 &   0.807$\pm$0.192 &                     0.801$\pm$0.189 &  0.797$\pm$0.207 \\ \cline{1-10} 
			\multicolumn{1}{c|}{\multirow{4}{*}{30}} & \multicolumn{1}{c|}{MAE} & 0.051$\pm$0.013 &                     \cellcolor{blue!8} 0.048$\pm$0.015 &  \cellcolor[RGB]{255,235,245} \textbf{0.046$\pm$0.015} &                                        0.049$\pm$0.016 &  0.092$\pm$0.055 &   0.066$\pm$0.023 &                     0.090$\pm$0.019 &               0.071$\pm$0.022 \\ 
			\multicolumn{1}{c|}{} & \multicolumn{1}{c|}{RMSE} & \cellcolor{blue!8} 0.076$\pm$0.021 &  \cellcolor[RGB]{255,235,245} \textbf{0.074$\pm$0.021} &  \cellcolor[RGB]{255,235,245} \textbf{0.074$\pm$0.021} &                     \cellcolor{blue!8} 0.076$\pm$0.023 &  0.175$\pm$0.065 &   0.121$\pm$0.041 &                     0.133$\pm$0.024 &               0.123$\pm$0.038 \\ 
			\multicolumn{1}{c|}{} & \multicolumn{1}{c|}{R2} & \cellcolor{blue!8} 0.913$\pm$0.061 &                                        0.912$\pm$0.060 &  \cellcolor[RGB]{255,235,245} \textbf{0.914$\pm$0.063} &                                        0.907$\pm$0.068 &  0.205$\pm$5.614 &   0.726$\pm$0.172 &                     0.728$\pm$0.103 &               0.740$\pm$0.131 \\ 
			\multicolumn{1}{c|}{} & \multicolumn{1}{c|}{CC} & \cellcolor{blue!8} 0.770$\pm$0.277 &                                        0.766$\pm$0.274 &  \cellcolor[RGB]{255,235,245} \textbf{0.779$\pm$0.269} &                                        0.760$\pm$0.284 &  0.683$\pm$0.270 &   0.740$\pm$0.261 &                     0.717$\pm$0.246 &  0.723$\pm$0.273 \\ 
			
\midrule
			
\multicolumn{10}{c}{{MOVER-SIS}} \\ 
\midrule
			\multicolumn{1}{c|}{\multirow{4}{*}{1}} & \multicolumn{1}{c|}{MAE} & \cellcolor[RGB]{255,235,245} \textbf{0.026$\pm$0.020} &                     \cellcolor{blue!8} 0.028$\pm$0.016 &                                        0.035$\pm$0.040 &                                        0.031$\pm$0.016 &  0.034$\pm$0.041 &   0.041$\pm$0.072 &                     0.044$\pm$0.070 &               0.044$\pm$0.074 \\ 
			\multicolumn{1}{c|}{} & \multicolumn{1}{c|}{RMSE} & \cellcolor[RGB]{255,235,245} \textbf{0.061$\pm$0.028} &  \cellcolor[RGB]{255,235,245} \textbf{0.061$\pm$0.028} &                                        0.070$\pm$0.047 &                     \cellcolor{blue!8} 0.065$\pm$0.028 &  0.086$\pm$0.044 &   0.101$\pm$0.074 &                     0.095$\pm$0.069 &               0.097$\pm$0.073 \\ 
			\multicolumn{1}{c|}{} & \multicolumn{1}{c|}{R2} & \cellcolor{blue!8} 0.911$\pm$0.434 &  \cellcolor[RGB]{255,235,245} \textbf{0.926$\pm$0.318} &                                        0.886$\pm$0.632 &                                        0.889$\pm$0.610 &  0.902$\pm$0.412 &   0.826$\pm$0.925 &                     0.830$\pm$0.929 &               0.823$\pm$1.043 \\ 
			\multicolumn{1}{c|}{} & \multicolumn{1}{c|}{CC} & \cellcolor[RGB]{255,235,245} \textbf{0.914$\pm$0.109} &                                        0.888$\pm$0.147 &                                        0.866$\pm$0.194 &                                        0.879$\pm$0.147 &  0.844$\pm$0.226 &   0.879$\pm$0.127 &  \cellcolor{blue!8} 0.896$\pm$0.106 &  0.848$\pm$0.211 \\ \cline{1-10} 
			\multicolumn{1}{c|}{\multirow{4}{*}{3}} & \multicolumn{1}{c|}{MAE} & \cellcolor[RGB]{255,235,245} \textbf{0.032$\pm$0.020} &                     \cellcolor{blue!8} 0.035$\pm$0.019 &                                        0.038$\pm$0.031 &                                        0.036$\pm$0.018 &  0.044$\pm$0.030 &   0.047$\pm$0.071 &                     0.056$\pm$0.068 &               0.051$\pm$0.073 \\ 
			\multicolumn{1}{c|}{} & \multicolumn{1}{c|}{RMSE} & \cellcolor[RGB]{255,235,245} \textbf{0.071$\pm$0.030} &  \cellcolor[RGB]{255,235,245} \textbf{0.071$\pm$0.033} &                                        0.078$\pm$0.040 &                     \cellcolor{blue!8} 0.072$\pm$0.031 &  0.105$\pm$0.039 &   0.113$\pm$0.071 &                     0.112$\pm$0.067 &               0.114$\pm$0.072 \\ 
			\multicolumn{1}{c|}{} & \multicolumn{1}{c|}{R2} & \cellcolor{blue!8} 0.868$\pm$0.696 &  \cellcolor[RGB]{255,235,245} \textbf{0.896$\pm$0.450} &                                        0.842$\pm$0.913 &                                        0.855$\pm$0.850 &  0.854$\pm$0.633 &   0.745$\pm$1.573 &                     0.751$\pm$1.454 &               0.733$\pm$1.692 \\ 
			\multicolumn{1}{c|}{} & \multicolumn{1}{c|}{CC} & \cellcolor[RGB]{255,235,245} \textbf{0.879$\pm$0.143} &                                        0.854$\pm$0.167 &                                        0.841$\pm$0.197 &                     \cellcolor{blue!8} 0.855$\pm$0.165 &  0.807$\pm$0.237 &   0.854$\pm$0.137 &                     0.836$\pm$0.161 &  0.821$\pm$0.224 \\ \cline{1-10} 
			\multicolumn{1}{c|}{\multirow{4}{*}{10}} & \multicolumn{1}{c|}{MAE} & \cellcolor[RGB]{255,235,245} \textbf{0.044$\pm$0.023} &                                        0.050$\pm$0.024 &                                        0.049$\pm$0.033 &                     \cellcolor{blue!8} 0.046$\pm$0.022 &  0.062$\pm$0.029 &   0.062$\pm$0.072 &                     0.081$\pm$0.063 &               0.073$\pm$0.061 \\ 
			\multicolumn{1}{c|}{} & \multicolumn{1}{c|}{RMSE} & \cellcolor{blue!8} 0.087$\pm$0.034 &                                        0.088$\pm$0.039 &                                        0.093$\pm$0.043 &  \cellcolor[RGB]{255,235,245} \textbf{0.085$\pm$0.036} &  0.138$\pm$0.042 &   0.144$\pm$0.073 &                     0.141$\pm$0.064 &               0.143$\pm$0.064 \\ 
			\multicolumn{1}{c|}{} & \multicolumn{1}{c|}{R2} & \cellcolor{blue!8} 0.785$\pm$1.303 &  \cellcolor[RGB]{255,235,245} \textbf{0.833$\pm$0.740} &                                        0.764$\pm$1.422 &                                        0.783$\pm$1.402 &  0.747$\pm$1.128 &   0.506$\pm$3.630 &                     0.576$\pm$2.732 &               0.567$\pm$2.966 \\ 
			\multicolumn{1}{c|}{} & \multicolumn{1}{c|}{CC} & \cellcolor[RGB]{255,235,245} \textbf{0.820$\pm$0.199} &                                        0.769$\pm$0.234 &                                        0.789$\pm$0.225 &                     \cellcolor{blue!8} 0.805$\pm$0.208 &  0.748$\pm$0.252 &   0.790$\pm$0.190 &                     0.742$\pm$0.224 &  0.742$\pm$0.256 \\ \cline{1-10} 
			\multicolumn{1}{c|}{\multirow{4}{*}{30}} & \multicolumn{1}{c|}{MAE} & \cellcolor{blue!8} 0.058$\pm$0.027 &                                        0.064$\pm$0.034 &                                        0.061$\pm$0.033 &  \cellcolor[RGB]{255,235,245} \textbf{0.056$\pm$0.028} &  0.091$\pm$0.041 &   0.087$\pm$0.067 &                     0.114$\pm$0.059 &               0.098$\pm$0.059 \\ 
			\multicolumn{1}{c|}{} & \multicolumn{1}{c|}{RMSE} & \cellcolor{blue!8} 0.101$\pm$0.037 &                                        0.104$\pm$0.050 &                                        0.106$\pm$0.042 &  \cellcolor[RGB]{255,235,245} \textbf{0.094$\pm$0.041} &  0.178$\pm$0.055 &   0.188$\pm$0.078 &                     0.173$\pm$0.064 &               0.176$\pm$0.064 \\ 
			\multicolumn{1}{c|}{} & \multicolumn{1}{c|}{R2} & 0.658$\pm$2.188 &  \cellcolor[RGB]{255,235,245} \textbf{0.781$\pm$0.942} &                                        0.626$\pm$2.428 &                     \cellcolor{blue!8} 0.700$\pm$2.011 &  0.577$\pm$1.924 &  -0.125$\pm$9.063 &                     0.321$\pm$4.297 &               0.273$\pm$4.833 \\ 
			\multicolumn{1}{c|}{} & \multicolumn{1}{c|}{CC} & \cellcolor[RGB]{255,235,245} \textbf{0.760$\pm$0.243} &                                        0.699$\pm$0.265 &                                        0.729$\pm$0.272 &                     \cellcolor{blue!8} 0.753$\pm$0.243 &  0.661$\pm$0.276 &   0.693$\pm$0.245 &                     0.648$\pm$0.257 &  0.668$\pm$0.270 \\ 
			\bottomrule
			
\end{tabular}
	\end{adjustbox}
\end{table*} 
\begin{table*}[!ht]
	\fontsize{10pt}{12pt}\selectfont
	\centering
	\renewcommand{\arraystretch}{1.1}
	\caption{Full performance comparison of different models on the incomplete variables track. The historical window is set to 30 minutes, and the forecasting horizons are 1, 3, 10, and 30 minutes.
	}
	\label{tab:main_masked_track_full}
	\begin{adjustbox}{max width=0.78\linewidth}
		\begin{tabular}{@{}cccccccc@{}}
\cline{1-8}
			\multicolumn{2}{c}{{\diagbox[width=8em, height=3.9em]{Metrics}{Models}}}                                                   & {\begin{tabular}[c]{@{}c@{}}MambaTS\\ (2024)\end{tabular}} & {\begin{tabular}[c]{@{}c@{}}iTransformer\\ (2023)\end{tabular}} & {\begin{tabular}[c]{@{}c@{}}FourierGNN\\ (2023)\end{tabular}} & {\begin{tabular}[c]{@{}c@{}}PatchTST\\ (2023)\end{tabular}} &  {\begin{tabular}[c]{@{}c@{}}Dlinear\\ (2023)\end{tabular}} & {\begin{tabular}[c]{@{}c@{}}S4\\ (2022)\end{tabular}} \\ 
			\midrule
\multicolumn{8}{c}{{VitalDB}} \\ 
			\midrule
\multicolumn{1}{c|}{\multirow{4}{*}{1}} & \multicolumn{1}{c|}{MAE} & \cellcolor[RGB]{255,235,245} \textbf{0.020$\pm$0.007} &                                        0.029$\pm$0.008 &                                        0.029$\pm$0.009 &  \cellcolor{blue!8} 0.026$\pm$0.007 &                     0.029$\pm$0.007 &              0.031$\pm$0.008 \\ 
			\multicolumn{1}{c|}{} & \multicolumn{1}{c|}{RMSE} & \cellcolor[RGB]{255,235,245} \textbf{0.040$\pm$0.012} &                     \cellcolor{blue!8} 0.046$\pm$0.012 &                                        0.050$\pm$0.014 &                     0.052$\pm$0.012 &                     0.063$\pm$0.014 &              0.061$\pm$0.013 \\ 
			\multicolumn{1}{c|}{} & \multicolumn{1}{c|}{R2} & \cellcolor[RGB]{255,235,245} \textbf{0.946$\pm$0.085} &                     \cellcolor{blue!8} 0.928$\pm$0.113 &                                        0.916$\pm$0.142 &                     0.912$\pm$0.161 &                     0.874$\pm$0.196 &              0.888$\pm$0.171 \\ 
			\multicolumn{1}{c|}{} & \multicolumn{1}{c|}{CC} & \cellcolor[RGB]{255,235,245} \textbf{0.930$\pm$0.094} &                     \cellcolor{blue!8} 0.923$\pm$0.099 &                                        0.885$\pm$0.134 &                     0.903$\pm$0.136 &                     0.902$\pm$0.109 &  0.885$\pm$0.144 \\ \cline{1-8} 
			\multicolumn{1}{c|}{\multirow{4}{*}{3}} & \multicolumn{1}{c|}{MAE} & \cellcolor[RGB]{255,235,245} \textbf{0.033$\pm$0.008} &                     \cellcolor{blue!8} 0.034$\pm$0.009 &                     \cellcolor{blue!8} 0.034$\pm$0.011 &  \cellcolor{blue!8} 0.034$\pm$0.009 &                     0.041$\pm$0.009 &              0.039$\pm$0.011 \\ 
			\multicolumn{1}{c|}{} & \multicolumn{1}{c|}{RMSE} & \cellcolor[RGB]{255,235,245} \textbf{0.054$\pm$0.014} &                     \cellcolor{blue!8} 0.055$\pm$0.013 &                                        0.057$\pm$0.016 &                     0.068$\pm$0.016 &                     0.079$\pm$0.017 &              0.076$\pm$0.017 \\ 
			\multicolumn{1}{c|}{} & \multicolumn{1}{c|}{R2} & \cellcolor[RGB]{255,235,245} \textbf{0.898$\pm$0.182} &                     \cellcolor{blue!8} 0.892$\pm$0.193 &                                        0.883$\pm$0.228 &                     0.847$\pm$0.316 &                     0.794$\pm$0.369 &              0.802$\pm$0.429 \\ 
			\multicolumn{1}{c|}{} & \multicolumn{1}{c|}{CC} & \cellcolor{blue!8} 0.869$\pm$0.148 &  \cellcolor[RGB]{255,235,245} \textbf{0.877$\pm$0.142} &                                        0.850$\pm$0.158 &                     0.864$\pm$0.152 &                     0.850$\pm$0.147 &  0.840$\pm$0.163 \\ \cline{1-8} 
			\multicolumn{1}{c|}{\multirow{4}{*}{10}} & \multicolumn{1}{c|}{MAE} & \cellcolor[RGB]{255,235,245} \textbf{0.037$\pm$0.013} &                     \cellcolor{blue!8} 0.039$\pm$0.012 &                                        0.042$\pm$0.013 &                     0.048$\pm$0.013 &                     0.060$\pm$0.012 &              0.055$\pm$0.014 \\ 
			\multicolumn{1}{c|}{} & \multicolumn{1}{c|}{RMSE} & \cellcolor[RGB]{255,235,245} \textbf{0.062$\pm$0.017} &                     \cellcolor{blue!8} 0.064$\pm$0.017 &                                        0.067$\pm$0.018 &                     0.091$\pm$0.022 &                     0.102$\pm$0.023 &              0.098$\pm$0.022 \\ 
			\multicolumn{1}{c|}{} & \multicolumn{1}{c|}{R2} & \cellcolor[RGB]{255,235,245} \textbf{0.848$\pm$0.339} &                     \cellcolor{blue!8} 0.839$\pm$0.341 &                                        0.821$\pm$0.402 &                     0.719$\pm$0.510 &                     0.657$\pm$0.561 &              0.694$\pm$0.499 \\ 
			\multicolumn{1}{c|}{} & \multicolumn{1}{c|}{CC} & \cellcolor[RGB]{255,235,245} \textbf{0.822$\pm$0.180} &                     \cellcolor{blue!8} 0.800$\pm$0.209 &                                        0.783$\pm$0.214 &                     0.786$\pm$0.195 &                     0.761$\pm$0.202 &  0.770$\pm$0.205 \\ \cline{1-8} 
			\multicolumn{1}{c|}{\multirow{4}{*}{30}} & \multicolumn{1}{c|}{MAE} & \cellcolor[RGB]{255,235,245} \textbf{0.047$\pm$0.016} &                     \cellcolor{blue!8} 0.050$\pm$0.016 &                                        0.051$\pm$0.016 &                     0.065$\pm$0.019 &                     0.082$\pm$0.017 &              0.071$\pm$0.022 \\ 
			\multicolumn{1}{c|}{} & \multicolumn{1}{c|}{RMSE} & \cellcolor[RGB]{255,235,245} \textbf{0.074$\pm$0.021} &                     \cellcolor{blue!8} 0.077$\pm$0.022 &                     \cellcolor{blue!8} 0.077$\pm$0.020 &                     0.114$\pm$0.032 &                     0.125$\pm$0.032 &              0.119$\pm$0.032 \\ 
			\multicolumn{1}{c|}{} & \multicolumn{1}{c|}{R2} & \cellcolor[RGB]{255,235,245} \textbf{0.765$\pm$0.495} &                     \cellcolor{blue!8} 0.749$\pm$0.507 &                                        0.745$\pm$0.539 &                     0.569$\pm$0.623 &                     0.490$\pm$0.661 &              0.550$\pm$0.604 \\ 
			\multicolumn{1}{c|}{} & \multicolumn{1}{c|}{CC} & \cellcolor[RGB]{255,235,245} \textbf{0.734$\pm$0.250} &                     \cellcolor{blue!8} 0.716$\pm$0.275 &                                        0.707$\pm$0.276 &                     0.689$\pm$0.251 &                     0.660$\pm$0.252 &  0.679$\pm$0.256 \\ 
			
\midrule
			
\multicolumn{8}{c}{{MOVER-SIS}} \\ 
			\midrule
			\multicolumn{1}{c|}{\multirow{4}{*}{1}} & \multicolumn{1}{c|}{MAE} & \cellcolor[RGB]{255,235,245} \textbf{0.032$\pm$0.015} &                                        0.041$\pm$0.022 &                                        0.041$\pm$0.017 &  \cellcolor{blue!8} 0.038$\pm$0.024 &                     0.048$\pm$0.029 &              0.042$\pm$0.027 \\ 
			\multicolumn{1}{c|}{} & \multicolumn{1}{c|}{RMSE} & \cellcolor[RGB]{255,235,245} \textbf{0.079$\pm$0.024} &                     \cellcolor{blue!8} 0.086$\pm$0.032 &                                        0.088$\pm$0.026 &                     0.092$\pm$0.035 &                     0.097$\pm$0.035 &              0.095$\pm$0.035 \\ 
			\multicolumn{1}{c|}{} & \multicolumn{1}{c|}{R2} & \cellcolor[RGB]{255,235,245} \textbf{0.901$\pm$0.214} &                                        0.890$\pm$0.237 &                     \cellcolor{blue!8} 0.892$\pm$0.211 &                     0.880$\pm$0.225 &                     0.868$\pm$0.243 &              0.873$\pm$0.239 \\ 
			\multicolumn{1}{c|}{} & \multicolumn{1}{c|}{CC} & \cellcolor[RGB]{255,235,245} \textbf{0.901$\pm$0.102} &                                        0.875$\pm$0.166 &                                        0.875$\pm$0.130 &                     0.873$\pm$0.154 &  \cellcolor{blue!8} 0.880$\pm$0.100 &  0.866$\pm$0.157 \\ \cline{1-8} 
			\multicolumn{1}{c|}{\multirow{4}{*}{3}} & \multicolumn{1}{c|}{MAE} & \cellcolor[RGB]{255,235,245} \textbf{0.039$\pm$0.017} &                     \cellcolor{blue!8} 0.044$\pm$0.023 &                                        0.048$\pm$0.020 &                     0.048$\pm$0.035 &                     0.057$\pm$0.035 &              0.051$\pm$0.037 \\ 
			\multicolumn{1}{c|}{} & \multicolumn{1}{c|}{RMSE} & \cellcolor[RGB]{255,235,245} \textbf{0.089$\pm$0.025} &                     \cellcolor{blue!8} 0.093$\pm$0.030 &                                        0.098$\pm$0.028 &                     0.107$\pm$0.041 &                     0.112$\pm$0.039 &              0.112$\pm$0.042 \\ 
			\multicolumn{1}{c|}{} & \multicolumn{1}{c|}{R2} & \cellcolor[RGB]{255,235,245} \textbf{0.874$\pm$0.280} &                                        0.863$\pm$0.337 &                     \cellcolor{blue!8} 0.864$\pm$0.293 &                     0.837$\pm$0.308 &                     0.826$\pm$0.319 &              0.825$\pm$0.328 \\ 
			\multicolumn{1}{c|}{} & \multicolumn{1}{c|}{CC} & \cellcolor[RGB]{255,235,245} \textbf{0.872$\pm$0.124} &                     \cellcolor{blue!8} 0.850$\pm$0.175 &                     \cellcolor{blue!8} 0.850$\pm$0.141 &                     0.839$\pm$0.176 &                     0.841$\pm$0.127 &  0.831$\pm$0.172 \\ \cline{1-8} 
			\multicolumn{1}{c|}{\multirow{4}{*}{10}} & \multicolumn{1}{c|}{MAE} & \cellcolor[RGB]{255,235,245} \textbf{0.052$\pm$0.024} &                     \cellcolor{blue!8} 0.055$\pm$0.024 &                                        0.061$\pm$0.025 &                     0.065$\pm$0.039 &                     0.079$\pm$0.036 &              0.070$\pm$0.042 \\ 
			\multicolumn{1}{c|}{} & \multicolumn{1}{c|}{RMSE} & \cellcolor[RGB]{255,235,245} \textbf{0.105$\pm$0.031} &                     \cellcolor{blue!8} 0.108$\pm$0.032 &                                        0.113$\pm$0.033 &                     0.136$\pm$0.045 &                     0.137$\pm$0.041 &              0.142$\pm$0.047 \\ 
			\multicolumn{1}{c|}{} & \multicolumn{1}{c|}{R2} & \cellcolor[RGB]{255,235,245} \textbf{0.818$\pm$0.511} &                                        0.802$\pm$0.707 &                     \cellcolor{blue!8} 0.816$\pm$0.411 &                     0.730$\pm$0.533 &                     0.728$\pm$0.554 &              0.704$\pm$0.601 \\ 
			\multicolumn{1}{c|}{} & \multicolumn{1}{c|}{CC} & \cellcolor[RGB]{255,235,245} \textbf{0.814$\pm$0.170} &                     \cellcolor{blue!8} 0.798$\pm$0.198 &                                        0.796$\pm$0.173 &                     0.771$\pm$0.202 &                     0.764$\pm$0.174 &  0.753$\pm$0.213 \\ \cline{1-8} 
			\multicolumn{1}{c|}{\multirow{4}{*}{30}} & \multicolumn{1}{c|}{MAE} & \cellcolor[RGB]{255,235,245} \textbf{0.068$\pm$0.027} &  \cellcolor[RGB]{255,235,245} \textbf{0.068$\pm$0.028} &                     \cellcolor{blue!8} 0.077$\pm$0.034 &                     0.090$\pm$0.041 &                     0.107$\pm$0.037 &              0.098$\pm$0.039 \\ 
			\multicolumn{1}{c|}{} & \multicolumn{1}{c|}{RMSE} & \cellcolor[RGB]{255,235,245} \textbf{0.117$\pm$0.036} &                     \cellcolor{blue!8} 0.120$\pm$0.037 &                                        0.127$\pm$0.041 &                     0.167$\pm$0.050 &                     0.161$\pm$0.044 &              0.171$\pm$0.047 \\ 
			\multicolumn{1}{c|}{} & \multicolumn{1}{c|}{R2} & \cellcolor[RGB]{255,235,245} \textbf{0.745$\pm$0.863} &                     \cellcolor{blue!8} 0.726$\pm$1.141 &  \cellcolor[RGB]{255,235,245} \textbf{0.745$\pm$0.621} &                     0.542$\pm$1.006 &                     0.578$\pm$0.949 &              0.516$\pm$1.122 \\ 
			\multicolumn{1}{c|}{} & \multicolumn{1}{c|}{CC} & \cellcolor[RGB]{255,235,245} \textbf{0.739$\pm$0.229} &                     \cellcolor{blue!8} 0.727$\pm$0.245 &                                        0.711$\pm$0.234 &                     0.666$\pm$0.245 &                     0.666$\pm$0.222 &  0.644$\pm$0.255 \\ 
\bottomrule
\end{tabular}
	\end{adjustbox}
\end{table*} 
\begin{table*}[!ht]
	\fontsize{10pt}{12pt}\selectfont
	\centering
	\renewcommand{\arraystretch}{1.1}
	\caption{Full performance comparison of different models on the cross-center generalization track. The historical window is set to 30 minutes, and the forecasting horizons are 1, 3, 10, and 30 minutes.
	}
	\label{tab:main_dg_track_full}
	\begin{adjustbox}{max width=0.9\linewidth}
		\begin{tabular}{@{}ccccccccc@{}}
\cline{1-9}
			\multicolumn{2}{c}{{\diagbox[width=8em, height=3.9em]{Metrics}{Models}}}                                                   & {\begin{tabular}[c]{@{}c@{}}\makecell[c]{Upper\\Bound} \\ \end{tabular}} & {\begin{tabular}[c]{@{}c@{}}MambaTS \\ (2024)\end{tabular}} & {\begin{tabular}[c]{@{}c@{}}iTransformer \\ (2023)\end{tabular}} & {\begin{tabular}[c]{@{}c@{}}FourierGNN \\ (2023)\end{tabular}} & {\begin{tabular}[c]{@{}c@{}}PatchTST \\ (2023)\end{tabular}} &  {\begin{tabular}[c]{@{}c@{}}Dlinear \\ (2023)\end{tabular}} & {\begin{tabular}[c]{@{}c@{}}S4 \\ (2022)\end{tabular}} \\ 
			\midrule
\multicolumn{9}{c}{{VitalDB $\to$ MOVER-SIS}} \\ 
			\midrule
\multicolumn{1}{c|}{\multirow{4}{*}{1}} & \multicolumn{1}{c|}{MAE} & \cellcolor[RGB]{255,235,245} \textbf{0.032$\pm$0.015} &                     \cellcolor{blue!8} 0.035$\pm$0.015 &                     0.053$\pm$0.024 &  0.062$\pm$0.024 &  0.042$\pm$0.025 &   0.060$\pm$0.028 &              0.048$\pm$0.028 \\ 
			\multicolumn{1}{c|}{} & \multicolumn{1}{c|}{RMSE} & \cellcolor[RGB]{255,235,245} \textbf{0.079$\pm$0.024} &                     \cellcolor{blue!8} 0.083$\pm$0.027 &                     0.100$\pm$0.034 &  0.113$\pm$0.034 &  0.098$\pm$0.036 &   0.101$\pm$0.032 &              0.103$\pm$0.036 \\ 
			\multicolumn{1}{c|}{} & \multicolumn{1}{c|}{R2} & \cellcolor[RGB]{255,235,245} \textbf{0.901$\pm$0.214} &                     \cellcolor{blue!8} 0.885$\pm$0.300 &                     0.839$\pm$0.502 &  0.797$\pm$0.524 &  0.865$\pm$0.258 &   0.853$\pm$0.340 &              0.854$\pm$0.265 \\ 
			\multicolumn{1}{c|}{} & \multicolumn{1}{c|}{CC} & \cellcolor[RGB]{255,235,245} \textbf{0.901$\pm$0.102} &                     \cellcolor{blue!8} 0.892$\pm$0.113 &                     0.850$\pm$0.141 &  0.805$\pm$0.174 &  0.862$\pm$0.155 &   0.864$\pm$0.151 &  0.846$\pm$0.164 \\ \cline{1-9} 
			\multicolumn{1}{c|}{\multirow{4}{*}{3}} & \multicolumn{1}{c|}{MAE} & \cellcolor[RGB]{255,235,245} \textbf{0.039$\pm$0.017} &                     \cellcolor{blue!8} 0.043$\pm$0.018 &                     0.058$\pm$0.026 &  0.061$\pm$0.026 &  0.051$\pm$0.037 &   0.062$\pm$0.034 &              0.054$\pm$0.038 \\ 
			\multicolumn{1}{c|}{} & \multicolumn{1}{c|}{RMSE} & \cellcolor[RGB]{255,235,245} \textbf{0.089$\pm$0.025} &                     \cellcolor{blue!8} 0.094$\pm$0.029 &                     0.110$\pm$0.036 &  0.116$\pm$0.036 &  0.115$\pm$0.043 &   0.113$\pm$0.038 &              0.117$\pm$0.043 \\ 
			\multicolumn{1}{c|}{} & \multicolumn{1}{c|}{R2} & \cellcolor[RGB]{255,235,245} \textbf{0.874$\pm$0.280} &                     \cellcolor{blue!8} 0.855$\pm$0.351 &                     0.807$\pm$0.572 &  0.789$\pm$0.515 &  0.817$\pm$0.327 &   0.821$\pm$0.354 &              0.813$\pm$0.329 \\ 
			\multicolumn{1}{c|}{} & \multicolumn{1}{c|}{CC} & \cellcolor[RGB]{255,235,245} \textbf{0.872$\pm$0.124} &                     \cellcolor{blue!8} 0.857$\pm$0.146 &                     0.824$\pm$0.152 &  0.794$\pm$0.179 &  0.831$\pm$0.160 &   0.836$\pm$0.145 &  0.821$\pm$0.173 \\ \cline{1-9} 
			\multicolumn{1}{c|}{\multirow{4}{*}{10}} & \multicolumn{1}{c|}{MAE} & \cellcolor[RGB]{255,235,245} \textbf{0.052$\pm$0.024} &                     \cellcolor{blue!8} 0.056$\pm$0.022 &                     0.066$\pm$0.026 &  0.073$\pm$0.028 &  0.068$\pm$0.045 &   0.080$\pm$0.036 &              0.071$\pm$0.045 \\ 
			\multicolumn{1}{c|}{} & \multicolumn{1}{c|}{RMSE} & \cellcolor[RGB]{255,235,245} \textbf{0.105$\pm$0.031} &                     \cellcolor{blue!8} 0.111$\pm$0.033 &                     0.123$\pm$0.038 &  0.128$\pm$0.039 &  0.145$\pm$0.050 &   0.138$\pm$0.041 &              0.148$\pm$0.050 \\ 
			\multicolumn{1}{c|}{} & \multicolumn{1}{c|}{R2} & \cellcolor[RGB]{255,235,245} \textbf{0.818$\pm$0.511} &                     \cellcolor{blue!8} 0.794$\pm$0.534 &                     0.745$\pm$0.867 &  0.726$\pm$0.795 &  0.696$\pm$0.576 &   0.721$\pm$0.570 &              0.688$\pm$0.574 \\ 
			\multicolumn{1}{c|}{} & \multicolumn{1}{c|}{CC} & \cellcolor[RGB]{255,235,245} \textbf{0.814$\pm$0.170} &                     \cellcolor{blue!8} 0.797$\pm$0.183 &                     0.764$\pm$0.195 &  0.734$\pm$0.218 &  0.758$\pm$0.199 &   0.755$\pm$0.191 &  0.748$\pm$0.205 \\ \cline{1-9} 
			\multicolumn{1}{c|}{\multirow{4}{*}{30}} & \multicolumn{1}{c|}{MAE} & \cellcolor[RGB]{255,235,245} \textbf{0.068$\pm$0.027} &                     \cellcolor{blue!8} 0.071$\pm$0.028 &                     0.076$\pm$0.031 &  0.082$\pm$0.033 &  0.093$\pm$0.050 &   0.108$\pm$0.038 &              0.094$\pm$0.051 \\ 
			\multicolumn{1}{c|}{} & \multicolumn{1}{c|}{RMSE} & \cellcolor[RGB]{255,235,245} \textbf{0.117$\pm$0.036} &                     \cellcolor{blue!8} 0.125$\pm$0.039 &                     0.131$\pm$0.041 &  0.136$\pm$0.044 &  0.181$\pm$0.061 &   0.164$\pm$0.045 &              0.184$\pm$0.062 \\ 
			\multicolumn{1}{c|}{} & \multicolumn{1}{c|}{R2} & \cellcolor[RGB]{255,235,245} \textbf{0.745$\pm$0.863} &                     \cellcolor{blue!8} 0.715$\pm$0.841 &                     0.692$\pm$0.954 &  0.666$\pm$0.945 &  0.472$\pm$1.076 &   0.564$\pm$0.974 &              0.466$\pm$1.036 \\ 
			\multicolumn{1}{c|}{} & \multicolumn{1}{c|}{CC} & \cellcolor[RGB]{255,235,245} \textbf{0.739$\pm$0.229} &                     \cellcolor{blue!8} 0.717$\pm$0.239 &                     0.694$\pm$0.246 &  0.674$\pm$0.256 &  0.656$\pm$0.248 &   0.661$\pm$0.232 &  0.652$\pm$0.249 \\ 
			
\midrule
			
\multicolumn{9}{c}{{MOVER-SIS $\to$ VitalDB}} \\ 
			\midrule
			
			\multicolumn{1}{c|}{\multirow{4}{*}{1}} & \multicolumn{1}{c|}{MAE} & \cellcolor[RGB]{255,235,245} \textbf{0.032$\pm$0.012} &                     \cellcolor{blue!8} 0.033$\pm$0.012 &                     0.040$\pm$0.011 &  0.041$\pm$0.014 &  0.038$\pm$0.012 &   0.046$\pm$0.011 &              0.042$\pm$0.013 \\ 
			\multicolumn{1}{c|}{} & \multicolumn{1}{c|}{RMSE} & \cellcolor[RGB]{255,235,245} \textbf{0.059$\pm$0.019} &                     \cellcolor{blue!8} 0.062$\pm$0.018 &                     0.072$\pm$0.018 &  0.071$\pm$0.020 &  0.079$\pm$0.019 &   0.089$\pm$0.019 &              0.083$\pm$0.019 \\ 
			\multicolumn{1}{c|}{} & \multicolumn{1}{c|}{R2} & \cellcolor[RGB]{255,235,245} \textbf{0.865$\pm$0.506} &                     \cellcolor{blue!8} 0.827$\pm$0.886 &                     0.713$\pm$1.919 &  0.776$\pm$1.131 &  0.618$\pm$2.811 &   0.562$\pm$2.922 &              0.550$\pm$3.441 \\ 
			\multicolumn{1}{c|}{} & \multicolumn{1}{c|}{CC} & \cellcolor[RGB]{255,235,245} \textbf{0.893$\pm$0.104} &  \cellcolor[RGB]{255,235,245} \textbf{0.893$\pm$0.098} &  \cellcolor{blue!8} 0.874$\pm$0.127 &  0.853$\pm$0.144 &  0.869$\pm$0.137 &   0.862$\pm$0.121 &  0.857$\pm$0.148 \\ \cline{1-9} 
			\multicolumn{1}{c|}{\multirow{4}{*}{3}} & \multicolumn{1}{c|}{MAE} & \cellcolor[RGB]{255,235,245} \textbf{0.039$\pm$0.014} &  \cellcolor[RGB]{255,235,245} \textbf{0.039$\pm$0.014} &  \cellcolor{blue!8} 0.043$\pm$0.013 &  0.046$\pm$0.016 &  0.047$\pm$0.014 &   0.054$\pm$0.013 &              0.050$\pm$0.015 \\ 
			\multicolumn{1}{c|}{} & \multicolumn{1}{c|}{RMSE} & \cellcolor[RGB]{255,235,245} \textbf{0.067$\pm$0.020} &                     \cellcolor{blue!8} 0.070$\pm$0.020 &                     0.075$\pm$0.020 &  0.078$\pm$0.021 &  0.093$\pm$0.022 &   0.101$\pm$0.021 &              0.098$\pm$0.022 \\ 
			\multicolumn{1}{c|}{} & \multicolumn{1}{c|}{R2} & \cellcolor[RGB]{255,235,245} \textbf{0.811$\pm$0.767} &                     \cellcolor{blue!8} 0.782$\pm$1.001 &                     0.657$\pm$2.314 &  0.717$\pm$1.420 &  0.455$\pm$3.894 &   0.369$\pm$4.446 &              0.342$\pm$5.027 \\ 
			\multicolumn{1}{c|}{} & \multicolumn{1}{c|}{CC} & \cellcolor[RGB]{255,235,245} \textbf{0.859$\pm$0.134} &  \cellcolor[RGB]{255,235,245} \textbf{0.859$\pm$0.126} &  \cellcolor{blue!8} 0.840$\pm$0.158 &  0.825$\pm$0.162 &  0.830$\pm$0.160 &   0.821$\pm$0.149 &  0.820$\pm$0.168 \\ \cline{1-9} 
			\multicolumn{1}{c|}{\multirow{4}{*}{10}} & \multicolumn{1}{c|}{MAE} & \cellcolor[RGB]{255,235,245} \textbf{0.050$\pm$0.018} &  \cellcolor[RGB]{255,235,245} \textbf{0.050$\pm$0.018} &  \cellcolor{blue!8} 0.054$\pm$0.016 &  0.057$\pm$0.018 &  0.063$\pm$0.017 &   0.073$\pm$0.016 &              0.067$\pm$0.018 \\ 
			\multicolumn{1}{c|}{} & \multicolumn{1}{c|}{RMSE} & \cellcolor[RGB]{255,235,245} \textbf{0.080$\pm$0.024} &                     \cellcolor{blue!8} 0.082$\pm$0.024 &                     0.088$\pm$0.024 &  0.090$\pm$0.025 &  0.117$\pm$0.027 &   0.122$\pm$0.026 &              0.123$\pm$0.029 \\ 
			\multicolumn{1}{c|}{} & \multicolumn{1}{c|}{R2} & \cellcolor[RGB]{255,235,245} \textbf{0.684$\pm$1.380} &                     \cellcolor{blue!8} 0.666$\pm$1.491 &                     0.466$\pm$3.705 &  0.493$\pm$3.103 &  0.325$\pm$3.500 &   0.066$\pm$6.455 &              0.147$\pm$5.193 \\ 
			\multicolumn{1}{c|}{} & \multicolumn{1}{c|}{CC} & \cellcolor[RGB]{255,235,245} \textbf{0.787$\pm$0.190} &  \cellcolor[RGB]{255,235,245} \textbf{0.787$\pm$0.186} &  \cellcolor{blue!8} 0.767$\pm$0.214 &  0.750$\pm$0.221 &  0.753$\pm$0.204 &   0.737$\pm$0.201 &  0.736$\pm$0.218 \\ \cline{1-9} 
			\multicolumn{1}{c|}{\multirow{4}{*}{30}} & \multicolumn{1}{c|}{MAE} & \cellcolor[RGB]{255,235,245} \textbf{0.062$\pm$0.022} &                     \cellcolor{blue!8} 0.063$\pm$0.023 &                     0.066$\pm$0.021 &  0.071$\pm$0.022 &  0.084$\pm$0.025 &   0.096$\pm$0.023 &              0.092$\pm$0.027 \\ 
			\multicolumn{1}{c|}{} & \multicolumn{1}{c|}{RMSE} & \cellcolor[RGB]{255,235,245} \textbf{0.093$\pm$0.029} &                     \cellcolor{blue!8} 0.095$\pm$0.030 &                     0.099$\pm$0.029 &  0.104$\pm$0.029 &  0.143$\pm$0.037 &   0.143$\pm$0.035 &              0.146$\pm$0.036 \\ 
			\multicolumn{1}{c|}{} & \multicolumn{1}{c|}{R2} & \cellcolor[RGB]{255,235,245} \textbf{0.531$\pm$2.009} &                     \cellcolor{blue!8} 0.523$\pm$2.042 &                     0.359$\pm$3.825 &  0.306$\pm$4.033 &  0.020$\pm$4.588 &  -0.205$\pm$7.647 &             -0.085$\pm$5.626 \\ 
			\multicolumn{1}{c|}{} & \multicolumn{1}{c|}{CC} & \cellcolor[RGB]{255,235,245} \textbf{0.697$\pm$0.261} &                     \cellcolor{blue!8} 0.694$\pm$0.260 &                     0.680$\pm$0.278 &  0.662$\pm$0.282 &  0.648$\pm$0.264 &   0.634$\pm$0.250 &  0.638$\pm$0.265 \\ 
			\bottomrule
			
\end{tabular}
	\end{adjustbox}
\end{table*}  
\end{document}